\documentclass{article}

\PassOptionsToPackage{numbers, compress}{natbib}
 \usepackage[preprint]{neurips_2026}

\usepackage[utf8]{inputenc} 
\usepackage[T1]{fontenc}    
\usepackage{hyperref}       
\hypersetup{
  colorlinks=true,
  citecolor={blue!60!black},
  linkcolor={red!60!black},
  urlcolor={blue!60!black},
}
\usepackage{url}            
\usepackage{booktabs}       
\usepackage{amsfonts}       
\usepackage{amsmath}        
\usepackage{nicefrac}       
\usepackage{microtype}      
\usepackage[table]{xcolor} 
\usepackage{graphicx}       
\usepackage{wrapfig}        
\usepackage{pifont}          
\usepackage{array}           
\usepackage{placeins}        
\usepackage{enumitem}        
\usepackage{algorithm}       
\usepackage{algpseudocode}   
\usepackage{amssymb}         
\usepackage[most]{tcolorbox}  
\usepackage{listings}        
\usepackage{graphicx}
\usepackage{hyperref}

\DeclareRobustCommand{\qustionicon}{%
  \raisebox{-0.4em}{\includegraphics[height=2.0em]{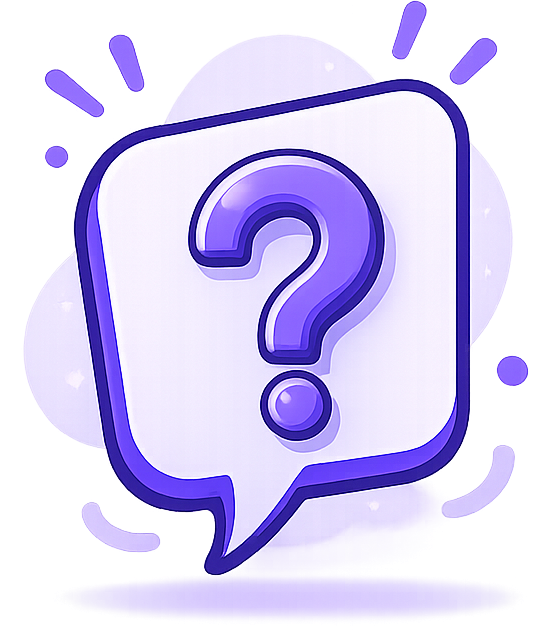}}\hspace{0.3em}%
}
\DeclareRobustCommand{\answericon}{%
  \raisebox{-0.3em}{\includegraphics[height=2.0em]{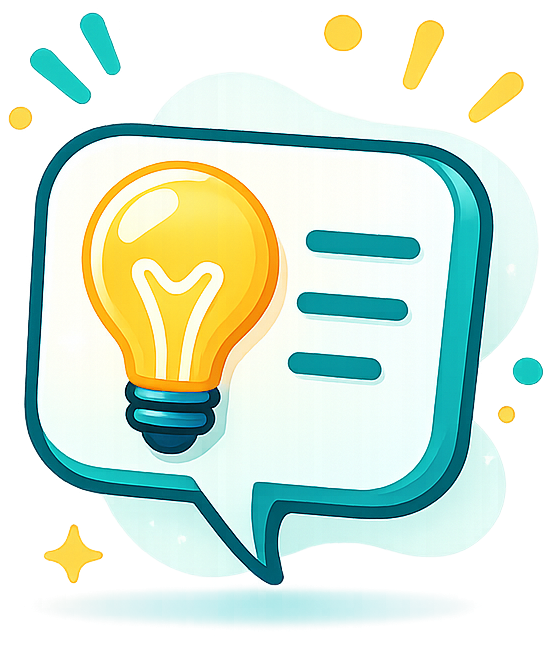}}\hspace{0.4em}%
}

\usepackage[font=small,labelfont=bf]{caption}  
\newtcolorbox{takeawaybox}{
  enhanced,
  colback=black!6,
  colframe=black!6,
  boxrule=0pt,
  arc=2pt,
  left=6pt, right=6pt, top=4pt, bottom=4pt,
  drop fuzzy shadow=black!50
}
\definecolor{dareblue}{HTML}{1a56db}
\newtcolorbox{designbox}[1][]{%
  enhanced, colback=blue!2, colframe=dareblue!60,
  boxrule=0.6pt, arc=3pt, left=6pt, right=6pt, top=2pt, bottom=2pt,
  before skip=4pt, after skip=4pt,
  fonttitle=\bfseries\small, coltitle=dareblue,
  attach boxed title to top left={yshift=-2mm, xshift=4mm},
  boxed title style={colback=white, colframe=white, boxrule=0pt},
  title={#1}
}
\newcommand{\icongear}{%
  \tikz[baseline=-0.05em, scale=0.85]{%
    \node[draw=dareblue, line width=0.7pt, circle, minimum size=0.7em,
          inner sep=0pt, font=\tiny\bfseries\color{dareblue}] {$\diamond$};
    \foreach \a in {0,60,...,300}{
      \draw[dareblue, line width=0.6pt] (\a:0.35em) -- (\a:0.5em);
    }
  }\xspace%
}
\usepackage{xspace}

\usepackage{xspace}

\lstdefinestyle{sparkprompt}{
  basicstyle=\ttfamily\scriptsize,
  breaklines=true,
  breakatwhitespace=false,
  columns=fullflexible,
  keepspaces=true,
  showstringspaces=false,
}
\newtcblisting{promptbox}[2][]{
  enhanced jigsaw,
  breakable,
  before skip=4pt,
  after skip=4pt,
  colback=black!3,
  colframe=black!15,
  boxrule=0.3pt,
  arc=2pt,
  left=4pt, right=4pt, top=2pt, bottom=2pt,
  title={#2},
  fonttitle=\bfseries\small,
  listing only,
  listing engine=listings,
  listing options={style=sparkprompt},
  #1
}

\newcommand{\appref}[1]{\hyperref[#1]{Appendix~\ref*{#1}}}

\makeatletter
\renewcommand\paragraph{\@startsection{paragraph}{4}{\z@}%
  {0.8ex \@plus 0.2ex \@minus 0.1ex}%
  {-1em}%
  {\normalfont\normalsize\bfseries}}
\makeatother

\newcommand{\cmark}{\textcolor{green!50!black}{\ding{51}}}
\newcommand{\xmark}{\textcolor{red}{\ding{55}}}

\usepackage{tikz}
\newcommand{\sknum}[1]{%
  \tikz[baseline=(c.base)]{%
    \node[circle,fill={blue!50!black},inner sep=0pt,minimum size=1.8ex,font=\scriptsize\bfseries\color{white}](c){#1};}%
}
\newcommand{\tknum}[1]{%
  \tikz[baseline=(c.base)]{%
    \node[circle,fill={violet!70!black},inner sep=0pt,minimum size=1.8ex,font=\scriptsize\bfseries\color{white}](c){#1};}%
}

\title{Evidence Over Plans: Online Trajectory Verification for Skill Distillation}

\author{%
  \begin{tabular}{c}
    {\bf Yang Zhou} \quad {\bf Zihan Dong} \quad {\bf Zhenting Wang} \\[2pt]
    {\bf Can Jin} \quad {\bf Shiyu Zhao} \quad {\bf Bangwei Guo} \quad {\bf Difei Gu} \\[2pt]
    {\bf Linjun Zhang} \quad {\bf Mu Zhou} \quad {\bf Dimitris N.~Metaxas} \\[4pt]
    Rutgers University \\[2pt]
    \texttt{eta.yang@rutgers.edu} \quad \texttt{dnm@rutgers.edu}
  \end{tabular}%
}

\begin{document}

\maketitle

\begin{abstract}
  Agent skills can remarkably improve task success rates by using human-written procedural documents, but their quality is difficult to assess without environment-grounded verification. Existing skill generation methods heavily rely on preference logs rather than direct environment interaction, often yielding negligible or even degraded gains. We identify that it is a fundamental timing bottleneck: robust skills should be \emph{posterior-based}, distilled from empirical environment interaction rather than prior plans. In this study, we introduce the \textbf{Posterior Distillation Index (PDI)}, a trajectory-level metric that quantifies how well a distilled skill is grounded in the task-environment evidence. To operationalize PDI, we present \textbf{SPARK} (\textbf{S}tructured \textbf{P}ipelines for \textbf{A}utonomous \textbf{R}unnable tas\textbf{K}s and s\textbf{K}ill generation) for preserving task execution evidence towards full trajectory-level analysis. SPARK generates environment-verified trajectories used to compute PDI, and it applies PDI as an online diagnostic and intervention signal to ensure posterior skill formation. Across 86 runnable tasks, SPARK-generated skills consistently surpass no-skill baselines and outperform human-written skills on student models (inference cost up to 1{,}000$\times$ cheaper than teacher models). These findings show that PDI-guided distillation produces efficient and transferable skills grounded in the task-environment interaction. We release our code in \href{https://github.com/EtaYang10th/spark-skills}{SPARK}.
\end{abstract}

\section{Introduction}
\label{sec:intro}

Large Language Model (LLM) agents are increasingly deployed on wide-ranging interactive software tasks~\citep{yao2023react,shinn2023reflexion,zhou2025led,zhou2025m,jin2024learning}. An emerging strategy is to equip agents with \emph{skills}~\citep{li2026skillsbench}, defined as in-context procedural documents that encode task-specific knowledge such as execution ordering, environment constraints, and verifier contracts. Benchmarks~\citep{li2026skillsbench,chen2026skillcraft} demonstrate that prior-based, human-written skills can remarkably boost task success rates, making such reusable procedural knowledge a promising abstraction for building scalable agent systems.

Yet a pessimistic picture appears that the quality of these skills is nearly impossible to assess without environment-grounded verification~\citep{li2026skillsbench}. Researchers propose lifelong skill accumulation~\citep{yang2026autoskill}, large-scale skill cataloging~\citep{liang2026skillnet}, or RL-based skill evolution~\citep{xia2026skillrl}. These studies lack trajectory verification and solely rely on preference logs. As a result, they can not produce transferable skills between teacher and student models at different capabilities. This raises a fundamental question: 

\noindent
\begin{center}
\textit{What makes skill distillation transferable across tasks and models efficiently and reliably?}
\end{center}

We identify that a key bottleneck is the timing of skill formation relative to the task-environment interaction. Existing skill generation~\citep{wang2023voyager,liang2026skillnet} asks the model to prescribe how a task should be solved before interacting with the environment, yielding skills dominated by generic priors and heuristics. Instead, we propose that effective procedural knowledge should be \emph{posterior-based} since its creation time: it must encode environment-specific constraints, execution dependencies, and failure modes discoverable only through environmental interactions. Once distilled, this posterior experience becomes reusable and verifiable guidance for future agent development.

\begin{figure}[t]
    \centering
    \includegraphics[width=\textwidth]{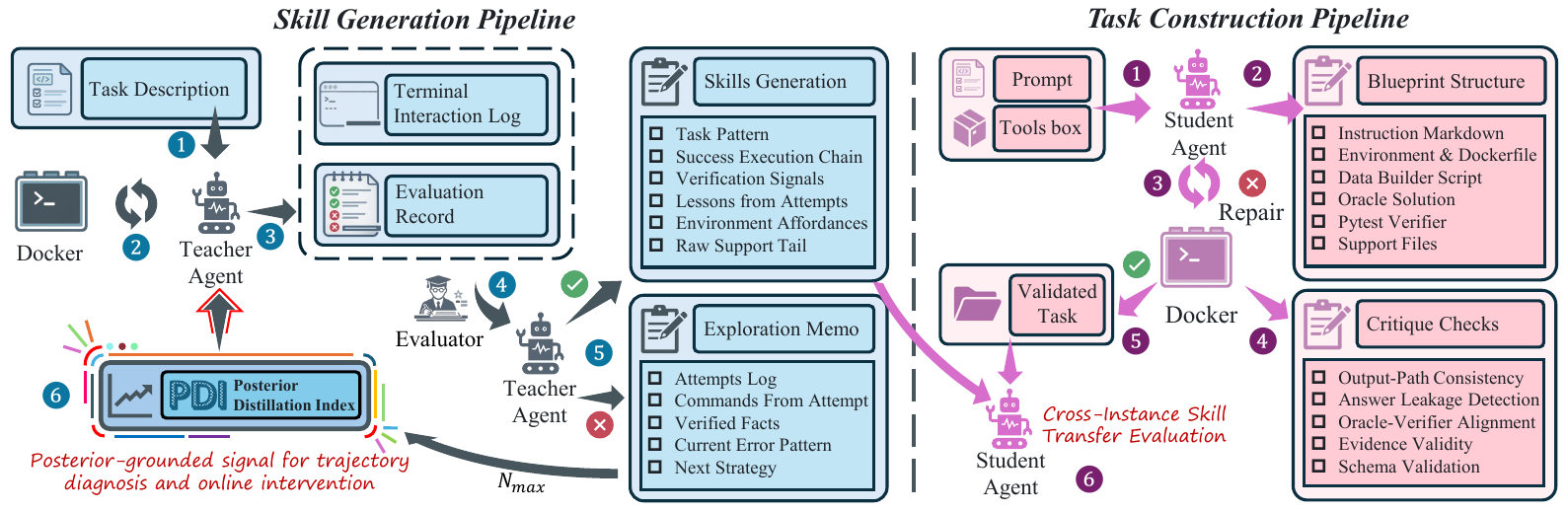}
    \caption{PDI-based SPARK Illustration. \textbf{1) Left (Skill Generation)}: Starting from a task description, a teacher agent (e.g., Claude Opus 4.6) interacts with a Dockerized environment (up to $N_{\max}$ attempts) and updates an exploration memo from execution feedback. Upon success, the full trajectory trace is distilled into \texttt{SKILL.md}. Upon failure, a PDI-based proxy triggers targeted interventions. \textbf{2) Right (Task Construction):} To validate the quality of skill generation, this is a supporting pipeline that converts prompt into oracle-verified benchmark instances via blueprint generation and critique checks. Then, a student agent (e.g., GPT 5.4 mini) evaluates the generated skill on these independently constructed tasks, testing whether PDI-grounded knowledge transfers across tasks rather than overfitting to the teacher's original trajectory.}
    \label{fig:pipeline}
    \vspace{-15pt}
\end{figure}

We introduce the \textbf{Posterior Distillation Index (PDI)}\footnote{`Posterior' denotes knowledge distilled from environment interaction. No probabilistic or Bayesian update is assumed.} as a principled metric that quantifies whether skills are grounded in environment-verified evidence rather than purely prior-based plans. By design, PDI aggregates three interpretable trajectory-level signals, including execution grounding, plan copying, and memo ossification, using an equal-weight linear combination. We deliberately avoid nonlinear fitting to ensure benchmark independence. To operationalize PDI, we present \textbf{SPARK} (\textbf{S}tructured \textbf{P}ipelines for \textbf{A}utonomous \textbf{R}unnable tas\textbf{K}s and s\textbf{K}ill generation). In~\autoref{fig:pipeline}, SPARK serves dual roles: it generates the environment-grounded trajectories from which PDI is computed; and it uses PDI as an online intervention signal to improve skill generation before its completion. Our main contributions are as follows.

\textbf{(1)}~\textbf{Trajectory-verified Metric}: PDI introduces a trajectory-level measure whether generated skills are grounded in task-environment evidence rather than unverified prior plans. PDI also provides a mechanistic assessment of which trajectories yield truly transferable skills (\autoref{sec:PDI}).

\textbf{(2)}~\textbf{Skill Generation}: SPARK makes skill generation transparent and analyzable by preserving execution logs, verifier signals, and memo histories towards full trajectory-level analysis (\autoref{sec:memo}).

\textbf{(3)}~\textbf{Online Intervention}: A memo-based proxy of PDI enables an online monitoring and intervention during skill exploration. Our primary \textbf{with/without PDI} ablation quantifies the performance gain enabled by this online intervention (\autoref{sec:pdi_online_control}).

\textbf{(4)}~\textbf{Trajectory Benchmark}: We provide a trajectory-level benchmark with full execution traces, verifier signals, memo histories, and distilled skills. This collected environment-verified evidence allows for evaluating skill transfer ability across environment interactions.

\FloatBarrier

\section{Related Work}
\label{sec:related}

\paragraph{Agent skill acquisition and evaluation.}
Agent skill acquisition has increasingly shifted from collecting experiences to distilling and deploying skills across tasks. Voyager~\citep{wang2023voyager} accumulates executable skills in an embodied environment. ExpeL~\citep{zhao2024expel} stores natural-language lessons from prior failures. Growing attempts distill the experience into transferable skill libraries, including AutoRefine~\citep{qiu2026autorefine}, Trace2Skill~\citep{ni2026trace2skill}, AutoSkill~\citep{yang2026autoskill}, and CUA-Skill~\citep{chen2026cuaskill}. SkillRouter~\citep{zheng2026skillrouter} further addresses the deployment by selecting highly-relevant skills from large registries. Beyond skill acquisition, skill evaluation is crucial to assess the quality of skills in real-world applications. SkillsBench~\citep{li2026skillsbench} evaluates when curated skills help at inference time. SkillWild~\citep{liu2026skillswild} studies more realistic retrieval-and-adaptation settings. Unlike these efforts, SPARK emphasizes interaction-grounded skill documents, cross-instance skill transfer, and trajectory-level analysis.
\vspace{-5pt}

\paragraph{LLM agents with external feedback.}
SPARK follows the interact--reflect--retry line of agentic system development, including ReAct~\citep{yao2023react}, Reflexion~\citep{shinn2023reflexion}, and ETO~\citep{song2024eto}. These studies mainly improve the decision making or trajectory revision within a single task-solving episode (one attempt at completing a given task). By contrast, we emphasize how agent execution traces can be distilled \citep{hinton2015distilling,furlanello2018born,jin2024learning} from the environment interaction that can transfer across tasks, turning trial-and-error records into a reusable knowledge base.
\vspace{-5pt}

\paragraph{Agent tooling and skill utility.}

LLM agent tooling and its transfer capability have evolved from learning tool policies to structuring skills. For instance, Toolformer~\citep{schick2023toolformer} and ToolLLM~\citep{qin2024toolllm} focus on learning to invoke tools. Meanwhile, CREATOR~\citep{qian2023creator} and LATM~\citep{cai2024latm} show that stronger models can synthesize reusable artifacts for weaker ones. Skill-oriented frameworks go further by packaging capabilities into modular, self-contained units with clear invocation protocols. SkillCraft~\citep{chen2026skillcraft} studies tool-usage skills, SkillNet~\citep{liang2026skillnet} organizes skills as reusable units, and SkillX~\citep{wang2026skillx} automatically constructs multi-level skill knowledge bases. In our study, SPARK highlights that the transferred artifact is a natural-language \texttt{SKILL.md} document rather than the executable code.

\section{Method}
\label{sec:method}

We study whether distilled skills are grounded in environment-verified evidence or merely reflect prior plans. To start, a teacher model generates skills from environment-interacted trajectories for use by a student model (here ``teacher'' denotes the behavior agent that produces exploration trajectories and ``student'' is the agent that consumes the learnable ones; no capacity asymmetry is implied). Our goal is to measure when these trajectories yield transferable procedural knowledge. We aim to define a trajectory-level score (i.e., Posterior Distillation Index (PDI)) that measures the degree to which the generated skill reflects posterior evidence rather than prior intent. In~\autoref{fig:pipeline}, SPARK produces major trajectories for PDI computation and uses PDI as an online signal to improve skill generation during the time of skill exploration. We describe skill generation (\autoref{sec:skill_gen}) and task construction pipeline (\autoref{sec:task_construction}) that enables this trajectory-level skill exploration.



\subsection{Skill Generation Pipeline}
\label{sec:skill_gen}

In \autoref{fig:pipeline} left, \sknum{1} the teacher agent interacts with a Dockerized environment up to $N_{\max}$ attempts. At each attempt, \sknum{2} teacher agent issues shell commands and observes terminal output. Then \sknum{3} teacher agent produces a \emph{Terminal Interaction Log} and \emph{Evaluation Record}. \sknum{4} When an attempt succeeds from the evaluator, \sknum{5} the full trajectory is distilled into a reusable document \texttt{SKILL.md} (\autoref{sec:evidence}); when it fails, the structured \emph{exploration memo} (\autoref{sec:memo}) is updated to capture the resulting evidence and revise the next strategy. This closed-loop design makes the generated skill explicitly traceable to the environment interaction. \sknum{6} Next, the Posterior Distillation Index (PDI, \autoref{sec:PDI}) is computed to quantify whether the accumulated trajectory is grounded in environment-verified evidence. Notably, PDI is not only a retrospective diagnostic signal since a proxy-based PDI can trigger intervention during skill exploration. This makes PDI useful both for trajectory analysis and improving skill generation online. Below we detail major blocks of evidence-driven skill generation and exploration memo. A full procedure is provided in \appref{app:alg_skill}.

    \subsubsection{Evidence-Driven Skill Generation}
    \label{sec:evidence}

    When a task is accomplished, SPARK assembles six structured evidence blocks to provide the skill-generation model with a comprehensive view of the solution path and obstacle encountered. This multi-source evidence design allows the generated skill toward reusable procedural structure rather than task-instance surface form, while establishing an explicit interface for evaluating predictive validity of individual evidence sources with respect to downstream skill utility.

    In \autoref{fig:pipeline} left:
    \textbf{1) Task Pattern} is a condensed task instruction capturing the problem class.
    \textbf{2) Execution Chain} offers key commands from the successful attempt, after semantic classification, importance scoring, and low-signal filtering.
    \textbf{3) Verification} records tests passed and the final reward.
    \textbf{4) Lessons} store repeated failure patterns and confirmed cautions distilled from the full attempt history.
    \textbf{5) Environment} offers runtime context from the Dockerfile (base image, packages, SDKs).
    \textbf{6) Raw Support Tail} shows the tail of the agent's stdout from the successful run.

    \subsubsection{Exploration Memo}
    \label{sec:memo}

    A single agent attempt can produce excessive amounts of tokens of raw stdout (standard output) and shell commands. Carrying this history forward verbatim would quickly exhaust the context window and dilute the model's attention. We therefore introduce the \emph{exploration memo} as a structured summary state that the teacher agent maintains across attempts. This design keeps the exploration process interpretable while preventing low-value repetition from overwhelming the context window.
    After each failed attempt, the teacher agent \emph{completely rewrites} the memo using recent agent's commands and a structured test summary, rather than appending to the old version. The memo follows a fixed five-section structure as shown in the left part of \autoref{fig:pipeline}:

    \begin{designbox}[\icongear~Memo Structure]
        \textbf{1) Attempts Log} stores a one-line LLM-distilled summary for each attempt; \textbf{2) Commands} record the key commands the agent actions in the most recent attempt; \textbf{3) Verified Facts} store the environment-confirmed outcomes across rewrites; \textbf{4) Current Error Pattern} identifies the active failure mode; and \textbf{5) Next Strategy} specifies an actionable plan that must differ from all previous approaches.
    \end{designbox}
    
    \noindent Notably, we preserve the full historical memo so that cross-attempt patterns remain available for subsequent analysis. SPARK's exploration-quality techniques are summarized in \appref{app:techniques}.
    
    \vspace{-8pt}
    

  \subsection{Task Construction Pipeline}
  \label{sec:task_construction}

In \autoref{fig:pipeline} right, \tknum{1} task-construction pipeline converts a prompt-level task idea into a structured task blueprint via three sequential stages. \tknum{2} A blueprint stage that exposes explicit intermediate structure; \tknum{3} a repair stage that iteratively fixes violated constraints in the blueprint; \tknum{4} a critique stage that checks semantic consistency and executable constraints; and \tknum{5} a validation stage that accepts only tasks passing deterministic oracle verification. Overall, task generation is a build-and-verify process that ensures each generated task is self-contained rather than a single-LLM-generated step. Further, \tknum{6} the pipeline supports cross-instance transfer evaluation by constructing unseen task variants from high-level specifications, on which a student agent is then deployed to test whether skills distilled from one trajectory generalize to related but independently constructed tasks.



\section{Experiments}
\label{sec:experiments}

We adopt the task suite from the SkillsBench benchmark~\citep{li2026skillsbench}, which provides 86 tasks spanning 11 domains (e.g., Software Engineering, Finance, and Cybersecurity), each paired with a natural-language instruction, a Docker execution environment, and a deterministic pytest oracle. To broaden our evaluation, SPARK (\autoref{sec:task_construction}) can synthesize runnable task variants from natural-language prompts; we use this capability to extend the evaluation pool (n = 300) and validate the generality power (\autoref{sec:ablation}). For cross-domain generalization, we evaluate SPARK on ALFWorld~\citep{shridhar2021alfworld}, a text-based household environment differing from terminal-based programming tasks. A capable teacher model interacts with the environment over up to $N_{\max}{=}7$ attempts; the objective is to distill the procedural knowledge embedded in these interactions into a reusable skill document that improves a weaker student model (statistics are reported in \appref{app:exploration_stats}).
The human-written skills used as a comparison baseline are those provided by SkillsBench~\citep{li2026skillsbench}; we use them as-is without modification.

\vspace{-3pt}
  \paragraph{Evaluation metrics.}
  Each task is associated with a deterministic pytest oracle that returns a reward $r \in [0,1]$, where $r{=}1$ indicates full task completion. We report two quantities throughout the study: \textbf{task reward} $r_{m,t}$, namely the reward obtained by student model $m$ on task $t$ under a given condition (baseline, with SPARK skill or with human-written skill); and \textbf{skill gain} $\Delta r_{m,t} = r_{m,t}^{\text{+skill}} - r_{m,t}^{\text{base}}$, namely the change in the reward attributable to the skill, where positive values indicate improvement and negative values indicate degradation.
  We aggregate across tasks via the mean reward $\bar{r}_m = \frac{1}{|\mathcal{T}|}\sum_{t} r_{m,t}$ and the mean skill gain $\overline{\Delta r}_m = \frac{1}{|\mathcal{T}|}\sum_{t} \Delta r_{m,t}$.

\vspace{-5pt}
\subsection{Main Results}
\label{sec:main_results}

\begin{figure}[t]
  \centering

  \begin{minipage}[t]{0.57\linewidth}
    \vspace{0pt}
    \centering
    \includegraphics[width=\linewidth]{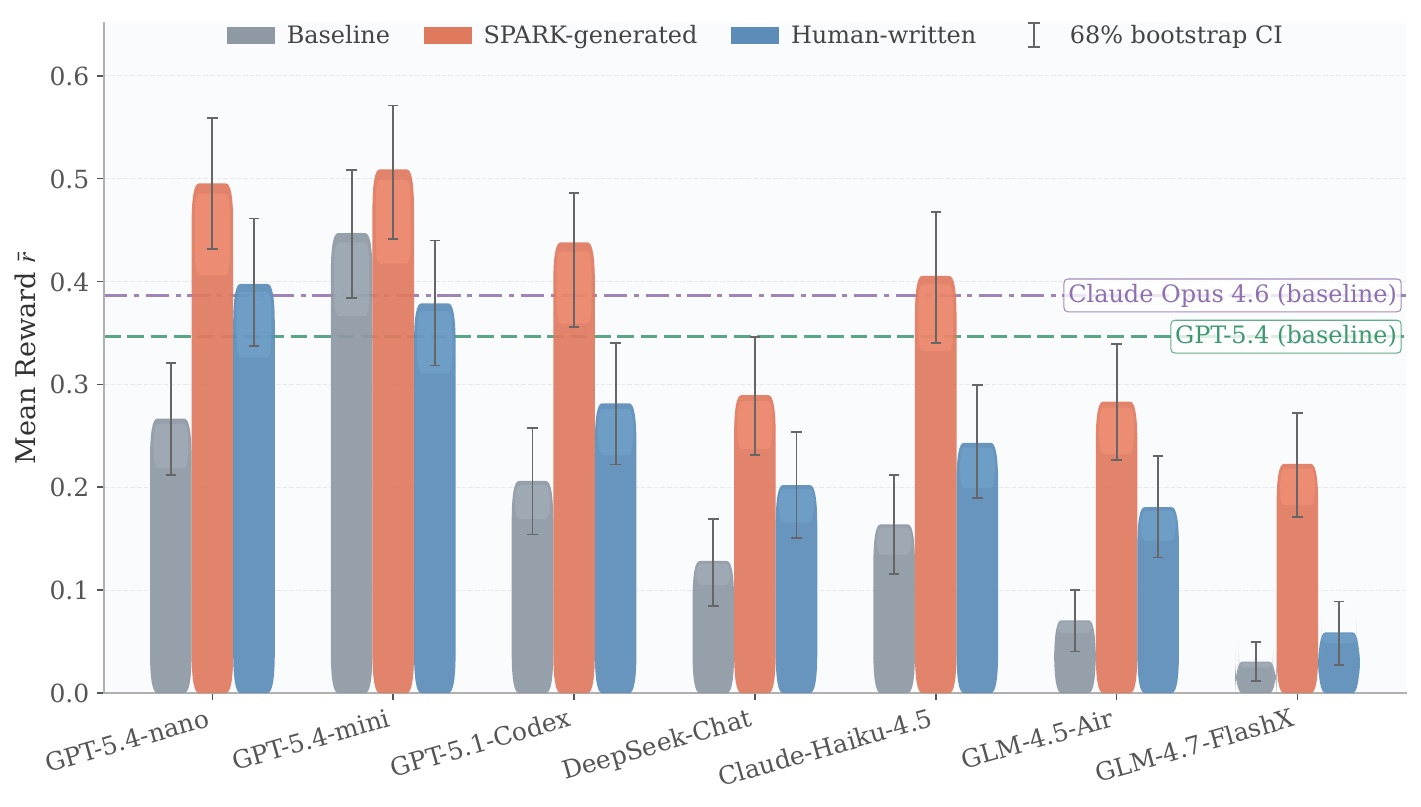}
    \caption{Mean reward $\bar{r}$ across seven student models under three conditions: no skill (baseline), SPARK-generated skills, and human-written skills. Horizontal dotted lines mark the interaction-free performance of two strong teacher models.}
    \label{fig:main_result}
  \end{minipage}
  \hfill
  \begin{minipage}[t]{0.405\linewidth}
    \vspace{0pt}
    \centering
    \includegraphics[width=\linewidth]{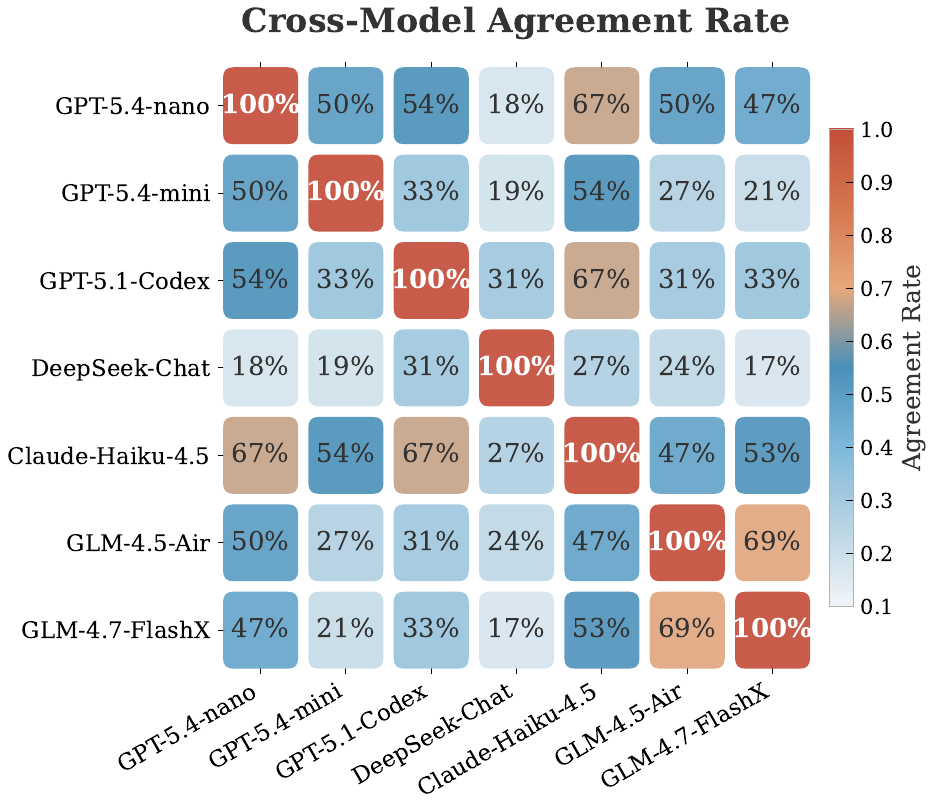}
    \caption{Cross-model agreement rate $A(m_i,m_j)$ among seven student models. Off-diagonal values are generally moderate.}
    \label{fig:agreement_heatmap}
  \end{minipage}
  \vspace{-8pt}
\end{figure}


\autoref{fig:main_result} reports the mean reward $\bar{r}$ of seven student models under three conditions: no skill (baseline), SPARK-generated skills, and human-written skills. Two horizontal lines mark the unaided performance of GPT-5.4 and Claude Opus 4.6 as the teacher models used during exploration.

Two striking findings stand out.
\textbf{First}, SPARK-generated skills consistently outperform human-written skills on the majority of student models: for instance, GPT-5.4-mini guided by SPARK skills reaches $\bar{r}{=}0.52$, surpassing the human-written condition ($0.47$).
\textbf{Second}, several small student models equipped with SPARK skills exceed the unaided performance of the teacher models themselves. For instance, GPT-5.4-nano with SPARK skills ($0.41$) outperforms Claude Opus 4.6 without skills ($0.37$). Overall, SPARK-distilled skills capture environment-specific insights that strong models alone miss, substantially narrowing the gap between weaker and stronger models. SPARK also carries a favorable cost profile: although strong-teacher exploration is expensive (\appref{app:exploration_stats}), the upfront cost amortizes across repeated cheap student runs---student inference costs as little as \$0.02 per task, over $1{,}000\times$ cheaper than teacher exploration---in some cases enabling weaker students to even surpass the teacher's own interaction-free performance.


\vspace{-8pt}
\subsection{Trajectory-Level Analysis}
\label{sec:trajectory_analysis}


Although SPARK-generated skills improve student performance (\autoref{sec:main_results}), they do not reveal \emph{why} some trajectories yield useful skills while others do not. Our central answer is the Posterior Distillation Index (PDI), which measures whether a distilled skill is grounded in posterior execution evidence rather than stale prior plans. Before formalizing PDI in \autoref{sec:PDI}, we first establish key motivating observations below (\appref{app:heatmap} details further trajectory correlation analysis).

%
\vspace{-8pt}
\subsubsection{\qustionicon Do a Few High-Quality Skills Drive All Gains?}
\label{sec:cross_model_disagreement}

\begin{takeawaybox}
\vspace{-4pt}
\emph{\answericon A handful of universally strong skills can not further improve skill utility with a moderate gain ($<$60\%), calling for more efforts on scale, divergence, and trajectory exploration.}
\end{takeawaybox}

For each pair $(m_i, m_j)$, we restrict to tasks where both fail the baseline ($r^{\mathrm{base}}_{m,t}{=}0$) and at least one shows a non-zero skill-induced change, and define the agreement rate as $A(m_i, m_j) = \frac{\bigl|\{t \mid \mathrm{sign}(\Delta r_{m_i,t}) = \mathrm{sign}(\Delta r_{m_j,t})\}\bigr|}{\bigl|\{t \mid \Delta r_{m_i,t} \neq 0 \lor \Delta r_{m_j,t} \neq 0\}\bigr|}$. \autoref{fig:agreement_heatmap} shows that off-diagonal rates are only moderate ($<$60\%). This calls for trajectory-level predictors of skill quality across models at different capacities.

\vspace{-8pt}
\subsubsection{\qustionicon Does More Evidence or More Attempts Improve Skills?}
\label{sec:compression_attempts}

\begin{takeawaybox}
\vspace{-4pt}
\emph{\answericon No, more evidence and attempts do not by themselves improve skills; excessive compression hurts and gains become unstable after the first a few rounds.}
\end{takeawaybox}

We next ask whether \emph{quantitative} factors alone, such as the volume of exploration evidence or the number of attempts, are sufficient to determine skill quality.

\paragraph{Compression ratio.}
Let $L_{\mathrm{traj}}$ denote the total character length of the agent's execution outputs (stdout) across all attempts, and $L_{\mathrm{skill}}$ the character length of the distilled \texttt{SKILL.md}. We define the compression ratio as $\rho_c = L_{\mathrm{traj}} / L_{\mathrm{skill}}$. \autoref{fig:three_panel}~(left) plots $\rho_c$ (log scale) against per-pair $\Delta$reward for all baseline-unsolved student$\times$task pairs. Across most student models, the Spearman correlation between $\rho_c$ and $\Delta$reward is negative, indicating that excessive compression discards actionable details and degrades skill effectiveness.

\paragraph{Exploration attempt budget.}
For each task, let $K$ denote the number of execution attempts the teacher performed before skill generation. For each student model $m$, we group all baseline-unsolved tasks by $K$ and compute the mean skill gain per bin as $\overline{\Delta r}_m(K) = \frac{1}{|\mathcal{T}_{m,K}|} \sum_{t \in \mathcal{T}_{m,K}} \Delta r_{m,t}$, where $\mathcal{T}_{m,K}$ is the set of tasks for which student model $m$ fails the baseline and the teacher uses exactly $K$ attempts. \autoref{fig:three_panel}~(middle) shows that gains are positive and relatively stable for $K \leq 3$, but become volatile and model-dependent beyond that point. More exploration therefore does not by itself explain why a distilled skill becomes useful.

\vspace{-8pt}
\subsubsection{\qustionicon Does Convergent or Divergent Exploration Produce Better Skills?}
\label{sec:conv_div}

\begin{takeawaybox}
\vspace{-4pt}
\emph{\answericon Divergent trajectories yield more transferable skills, whereas convergent ones encode teacher-specific refinements that weaker students cannot reproduce.}
\end{takeawaybox}

We next ask whether the instability above depends on \emph{how} the teacher explores, namely whether it stays within one line of attack or pivots to different ones.

\paragraph{Exploration-mode classification.}
Let $\mu_i$ denote the memo text after the $i$-th reflection. We measure the Jaccard similarity between successive memos as $J_i = \frac{|\,\mathrm{tok}(\mu_{i-1}) \cap \mathrm{tok}(\mu_i)\,|}{|\,\mathrm{tok}(\mu_{i-1}) \cup \mathrm{tok}(\mu_i)\,|}$, where $\mathrm{tok}(\cdot)$ denotes word-level tokenization. We then fit a linear trend $J_i \approx \alpha + \gamma \cdot i$ and use the slope $\gamma$ to classify each task: \textbf{Convergent} ($\gamma > 0$), where successive memos become increasingly similar, and \textbf{Divergent} ($\gamma \leq 0$), where memos grow less similar over time.

\begin{figure}[t]
  \centering
  \begin{minipage}[t]{0.32\textwidth}
    \centering
    \includegraphics[width=\linewidth]{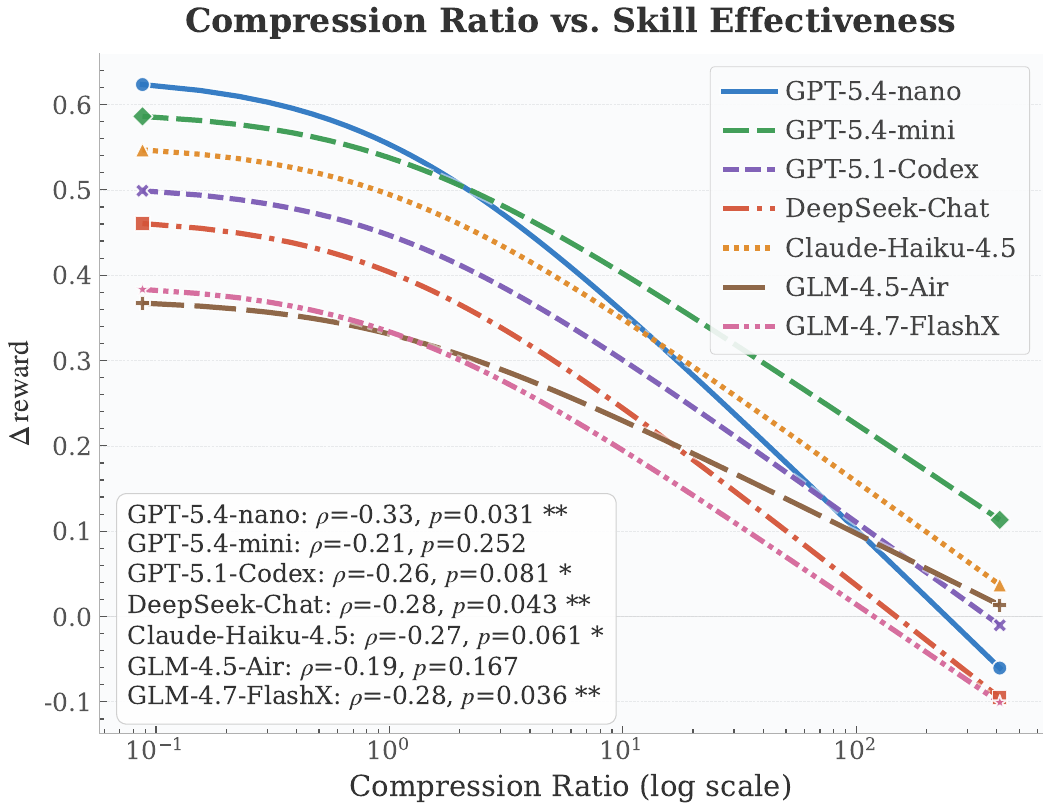}
  \end{minipage}%
  \hfill
  \begin{minipage}[t]{0.35\textwidth}
    \centering
    \includegraphics[width=\linewidth]{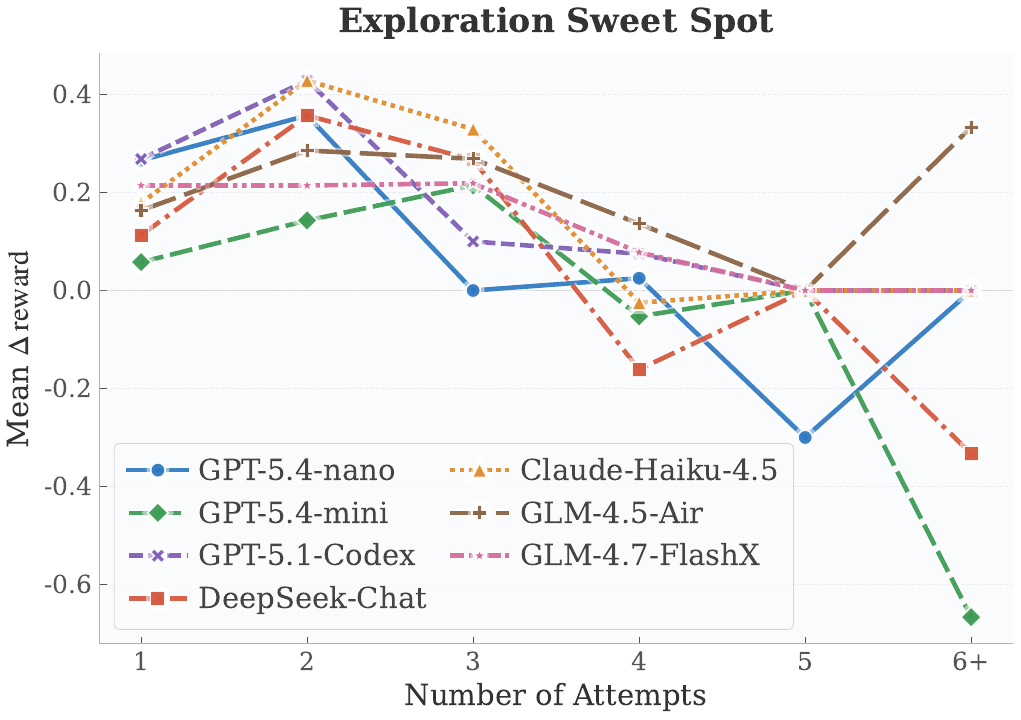}
  \end{minipage}%
  \hfill
  \begin{minipage}[t]{0.32\textwidth}
    \centering
    \includegraphics[width=\linewidth]{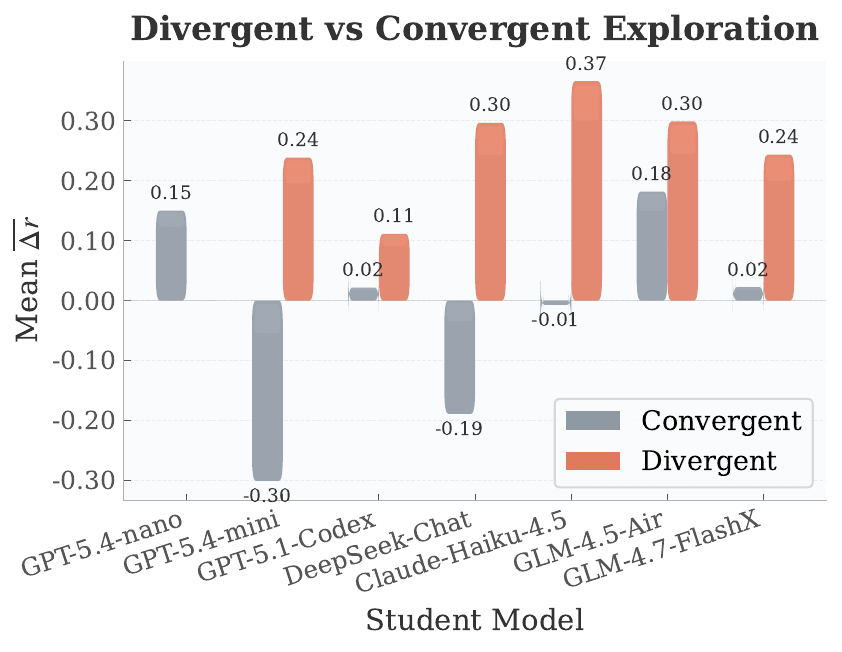}
  \end{minipage}
  \caption{Three complementary views of skill quality determinants. \textbf{Left:} Compression ratio $\rho_c$ vs.\ per-pair $\Delta r$; excessive compression degrades skill effectiveness. \textbf{Middle:} Mean $\Delta r$ as a function of the number of exploration attempts; gains are stable for the first three attempts and become volatile thereafter. \textbf{Right:} Mean $\overline{\Delta r}$ per student model for skills distilled from convergent vs.\ divergent teacher trajectories.}
  \vspace{-10pt}
  \label{fig:three_panel}
\end{figure}

\autoref{fig:three_panel}~(right) reports the mean skill gain $\overline{\Delta r}$ for each student model, grouped by exploration mode. Across nearly all student models, skills distilled from divergent trajectories yield substantially larger gains than those from convergent ones; for several models, convergent-trajectory skills even produce negative $\Delta r$ relative to the no-skill baseline.

This contrast reveals that exploration quality depends not only on search volume, but also on whether the trajectory accumulated genuinely new posterior evidence rather than refining the same plan.

\vspace{-8pt}
\subsubsection{\qustionicon What Separates Useful Iterative Skills from Useless Ones?}
\label{sec:prior_posterior}

\begin{takeawaybox}
\vspace{-4pt}
\emph{\answericon Environment-verified skills consistently outperform human-authored baselines, while plan-copying skills without external task-update signals fall below even interaction-free ones.}
\end{takeawaybox}

Prior analyses motivate us to hypothesize that useful skills emerge when the final document distills posterior execution evidence rather than echoing prior plans. We call this property the Posterior Distillation Index (PDI) (\autoref{sec:PDI}). To formalize it, we compare \emph{token-frequency distributions} between trajectory segments and the final skill to capture both token presence and its emphasis.

\paragraph{Token-distribution setup.}
For any text segment $x$ and a fixed vocabulary $\mathcal{V}$ constructed from the trajectory, let $c_x(w)$ denote the count of token $w$. We define the additively smoothed empirical distribution
$P_x(w) = \frac{c_x(w) + \alpha}{\sum_{v \in \mathcal{V}} c_x(v) + \alpha |\mathcal{V}|}, \qquad \alpha > 0$.
To measure how probability mass shifts between trajectory segments and the final skill, we need a discrepancy that is sensitive to token \emph{frequency}, not merely token \emph{presence}. The Kullback--Leibler (KL) divergence $\mathrm{KL}(P \,\|\, Q) = \sum_{w} P(w)\log \frac{P(w)}{Q(w)}$ is a natural candidate: it quantifies the coding inefficiency of using $Q$ to represent $P$. However, KL is asymmetric and unbounded, which complicates interpretation when neither distribution is a clear reference. We thus adopt the Jensen--Shannon divergence (JSD), a symmetrized and bounded variant:

\begin{equation}
  \mathrm{JS}(P_x, P_y)
  \;=\;
  \tfrac{1}{2}\,\mathrm{KL}(P_x \,\|\, M)
  \;+\;
  \tfrac{1}{2}\,\mathrm{KL}(P_y \,\|\, M),
  \qquad
  M = \tfrac{P_x + P_y}{2}\,,
  \label{eq:jsd}
\end{equation}

which lies in $[0,1]$ under base-2 logarithms. JSD inherits KL's sensitivity to how probability mass is allocated across tokens, while remaining symmetric and well-defined even for sparse distributions. We convert divergence to similarity via $\psi(P_x,P_y) = 1 - \mathrm{JS}(P_x,P_y)$, so that higher values indicate stronger distributional alignment. We examine three dynamic properties of each trajectory below:

\begin{itemize}[leftmargin=12pt]
    \item\textbf{Plan Copying} ($\phi_{\mathrm{plan}}$).
Let $P_P$ denote the token distribution of all accumulated \emph{Next Strategy} sections from the exploration memo, and $P_s$ the token distribution of the final \texttt{SKILL.md}. We define plan copying as $\phi_{\mathrm{plan}} = \psi(P_P, P_s)$. High $\phi_{\mathrm{plan}}$ indicates that the skill's token emphasis closely mirrors the memo's planned actions rather than encoding independently verified knowledge.
    
    \item\textbf{Execution Grounding} ($\phi_{\mathrm{exec}}$).
Let $P_E$ denote the distribution of agent commands from the successful execution trace. We define execution grounding as $\phi_{\mathrm{exec}} = \psi(P_E, P_s)$. High $\phi_{\mathrm{exec}}$ indicates that the skill concentrates on operations by operations validated by the environment.
    
    \item\textbf{Memo Ossification} ($\phi_{\mathrm{oss}}$).
We measure two cross-attempt stability signals: (i)~the distributional similarity of \emph{Verified Facts} between consecutive reflections, $\psi(P_{v_{i-1}}, P_{v_i})$, and (ii)~the distributional persistence of failed test sets, $\psi(P_{f_{i-1}}, P_{f_i})$. We define memo ossification as the equal-weight average $\phi_{\mathrm{oss}} = \frac{1}{2}\,\overline{\psi(P_{v_{i-1}}, P_{v_i})} + \frac{1}{2}\,\overline{\psi(P_{f_{i-1}}, P_{f_i})}$, where $\overline{(\cdot)}$ denotes the mean across consecutive attempt pairs. High $\phi_{\mathrm{oss}}$ indicates that the trajectory is stuck, repeating the same task understanding and failure patterns without substantive update.
\end{itemize}

\subsubsection{Posterior Distillation Index (PDI)}
\label{sec:PDI}

We aggregate above three characteristics into a single composite score. Let $z(\cdot)$ denote z-score normalization across all iterative tasks. The \emph{Posterior Distillation Index} (PDI) is defined as:
\begin{equation}
  \mathrm{PDI} \;=\; z(\phi_{\mathrm{exec}}) \;-\; z(\phi_{\mathrm{plan}}) \;-\; z(\phi_{\mathrm{oss}})\,.
  \label{eq:pdi}
\end{equation}
Intuitively, a high PDI indicates that the skill is distributionally aligned with environment-verified evidence (high $\phi_{\mathrm{exec}}$), diverges from memo-level plans (low $\phi_{\mathrm{plan}}$), and stems from a trajectory whose task understanding genuinely evolves rather than repeating the same failure pattern (low $\phi_{\mathrm{oss}}$). We intentionally use a simple equal-weight linear form because PDI is designed to be an \textit{interpretable} and \textit{transferable} diagnostic index rather than a benchmark-specific predictive model. Indeed, sign-consistent weight combinations yield significant correlations ($p{<}0.05$) with skill gain, and cross-validated weight fitting underperforms on held-out data (\appref{app:weight_sensitivity}), confirming that PDI's predictive power derives from its directional structure rather than any particular coefficient. Throughout the experiments, we instantiate PDI with $\mathcal{D}{=}\mathrm{JS}$ and a smoothing parameter $\alpha{=}0.002$; sensitivity to $\alpha$ is analyzed below.

\begin{figure}[t]
  \centering
  \includegraphics[width=\linewidth]{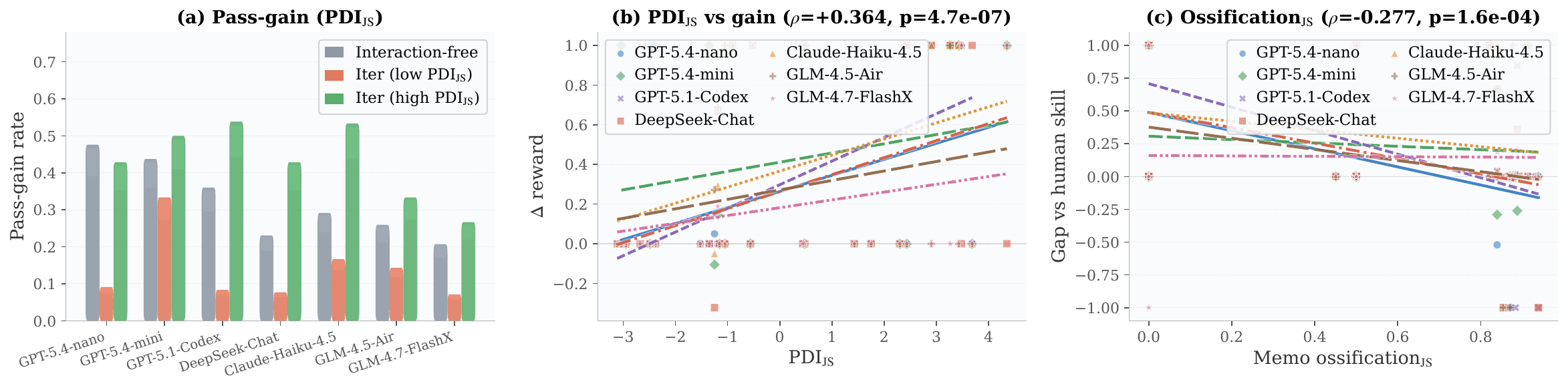}
  \caption{Trajectory-level analysis of skill quality using divergence-based PDI ($\alpha{=}0.002$). (a)~Pass-gain rate by trajectory group across seven student models: high-PDI iterative trajectories consistently outperform both interaction-free and low-PDI iterative skills. (b)~PDI vs.\ per-pair $\Delta r$ ($\rho{=}{+}0.364$, $p{<}10^{-6}$). (c)~Memo ossification vs.\ gap relative to human-written skills ($\rho{=}{-}0.277$, $p{<}10^{-3}$): trajectories that repeat stale task understanding produce skills that fall further behind human baselines.}
  \label{fig:pdi_analysis}
  \vspace{-10pt}
\end{figure}

\paragraph{PDI predicts student-level gains.}
We evaluate the predictive power of PDI through three complementary views (\autoref{fig:pdi_analysis}), each targeting a different facet of skill usefulness.

\emph{Panel~(a): Pass-gain rate.}
For each trajectory group
$g \in \{\text{interaction-free},\allowbreak\,\text{iter-low-PDI},\allowbreak\,\text{iter-high-PDI}\}$
and student model $m$, we compute the pass-gain rate over baseline-unsolved pairs as $\mathrm{PG}(g, m) = \frac{\bigl|\{t \in \mathcal{T}_g \mid r_{m,t}^{\text{base}}{=}0 \wedge r_{m,t}^{\text{+skill}}{=}1\}\bigr|}{\bigl|\{t \in \mathcal{T}_g \mid r_{m,t}^{\text{base}}{=}0\}\bigr|}$, where $\mathcal{T}_g$ denotes the set of tasks in trajectory group $g$. This quantity is the probability that a student passes with the skill given that it failed without one.
Skills from high-PDI iterative trajectories consistently achieve the highest pass-gain rates across all seven student models, outperforming both interaction-free skills and low-PDI iterative skills, the latter of which often fall below 0.10.

\emph{Panel~(b): PDI vs.\ skill gain.}
For each baseline-unsolved iterative pair $(m, t)$, we plot PDI against the skill gain $\Delta r_{m,t}$. The scatter reveals a significant positive Spearman correlation ($\rho{=}{+}0.364$, $p{=}4.7{\times}10^{-7}$): higher PDI is associated with larger student gains across student models.

\emph{Panel~(c): Memo ossification vs.\ gap to human skills.}
We define the gap relative to human-written skills as $G_{m,t} = \Delta r_{m,t}^{\text{SPARK}} - \Delta r_{m,t}^{\text{human}}$, where $\Delta r_{m,t}^{\text{SPARK}}$ and $\Delta r_{m,t}^{\text{human}}$ are the skill gains from SPARK-generated and human-written skills, respectively. Plotting $\phi_{\mathrm{oss}}$ against $G_{m,t}$ yields a negative correlation ($\rho{=}{-}0.277$, $p{=}1.6{\times}10^{-4}$), confirming that cognitive stagnation during exploration directly harms downstream transfer.

A Mann--Whitney U test between high- and low-PDI task groups yields $p{=}0.028$, confirming that PDI retains discriminative power at the task level.

\paragraph{Effective skills ground in evidence rather than plans.}

\begin{table}[t]
  \centering
  \begin{minipage}[t]{0.34\linewidth}
    \vspace{0pt}
    \centering
    \small
    \setlength{\tabcolsep}{4pt}
    \renewcommand{\arraystretch}{1.15}
    \captionof{table}{Mean $\Delta r$ and gap to human-written skills on tasks, grouped by plan copying ($\phi_{\mathrm{plan}}$) and execution grounding ($\phi_{\mathrm{exec}}$). High/Low denotes above/below median, and low $\phi_{\mathrm{plan}}$ with high $\phi_{\mathrm{exec}}$ performs best.}
    \label{tab:quadrant}
    \begin{tabular}{@{} l l c c @{}}
      \toprule
      \textbf{$\phi_{\mathrm{plan}}$} & \textbf{$\phi_{\mathrm{exec}}$} & \textbf{$\overline{\Delta r}$} & \textbf{Gap2Human}  \\
      \midrule
      Low  & High & $+$\textbf{0.377} & \textbf{$+$0.312} \\
      Low  & Low  & $+$0.321 & $+$0.393 \\
      High & High & $+$0.143 & $+$0.071  \\
      High & Low  & $+$0.028 & $-$0.040 \\
      \bottomrule
    \end{tabular}
    \vspace{2pt}
  \end{minipage}
  \hfill
  \begin{minipage}[t]{0.62\linewidth}
    \vspace{0pt}
    \centering
    \small
    \setlength{\tabcolsep}{2.5pt}
    \renewcommand{\arraystretch}{1.12}
    \captionof{table}{Ablation results on downstream pass rate and task-variant transfer. Left: pass rate on the PDI-rerun task subset under skills generated without and with PDI-guided online intervention. Right: transfer to 300 additional runnable task variants generated by the layered task-construction pipeline.}
    \label{tab:pdi_eval}
    \label{tab:variant_transfer}
    \resizebox{\linewidth}{!}{%
    \begin{tabular}{@{} l c c c !{\hspace{4pt}\vrule\hspace{4pt}} c c c c @{}}
      \toprule
      \multicolumn{4}{c}{\textbf{Pass rate}} & \multicolumn{4}{c}{\textbf{Task variants}} \\
      \cmidrule(lr){1-4} \cmidrule(lr){5-8}
      \textbf{Model} & \textbf{w/o PDI} & \textbf{w/ PDI} & \textbf{$\Delta$} & \textbf{Base} & \textbf{Gen.} & \textbf{Hum.} & \textbf{$\Delta$} \\
      \midrule
      DeepSeek-Chat & 9.1\% & 42.4\% & $+$33.3\% & 43\% & \textbf{87\%} & 80\% & \textbf{+43\%} \\
      GPT-5.1-Codex & 0.0\% & 24.2\% & $+$21.2\% & 17\% & \textbf{83\%} & 50\% & \textbf{+67\%} \\
      GPT-5.4-mini & 18.2\% & 36.4\% & $+$12.1\% & 7\% & \textbf{53\%} & 33\% & \textbf{+47\%} \\
      GPT-5.4-nano & 15.2\% & 30.3\% & $+$15.2\% & 56.7\% & \textbf{90.0\%} & 83.3\% & \textbf{+33.3\%} \\
      GLM-4.5-Air & 3.0\% & 15.2\% & $+$12.1\% & 45.0\% & \textbf{90.0\%} & 82.0\% & \textbf{+45.0\%} \\
      \bottomrule
    \end{tabular}%
    }
    \vspace{2pt}
  \end{minipage}
  \vspace{-25pt}
\end{table}


To disentangle the contributions of plan copying and execution grounding, we partition iterative tasks into four quadrants by the median of $\phi_{\mathrm{plan}}$ and $\phi_{\mathrm{exec}}$. \autoref{tab:quadrant} reports the mean $\Delta r$ and the mean gap versus human-written skills for each quadrant. The best-performing quadrant, low plan copying combined with high execution grounding, achieves a mean $\Delta r$ of $+$0.377 and a positive gap of $+$0.312 over human skills. In contrast, the worst quadrant, high plan copying with low execution grounding, yields a near-zero mean $\Delta r$ of $+$0.028 and a gap of $-$0.040 below human skills. This confirms that useful skills are not summaries of what the agent \emph{planned} to do, but distillations of what the environment \emph{verified} to work.

\paragraph{Belief stagnation undermines exploration.}
A related failure mode is what we term the \emph{facts--strategy gap}: the difference between the cross-attempt stability of \emph{Verified Facts} and \emph{Next Strategy}. A large gap indicates that the agent superficially varies its strategy across retries while its task understanding remains unchanged, a form of pseudo-exploration. Among iterative tasks, this gap is negatively correlated with mean $\Delta$reward ($\rho{=}{-}0.390$, $p{=}0.033$), confirming that trajectories in which ``strategies change but task understanding does not'' fail to produce transferable skills.

\vspace{-4pt}
\paragraph{Sensitivity to the smoothing parameter.}

PDI is robust to the choice of $\alpha$ within a broad low-smoothing regime; the full sensitivity curves and significance analysis are provided in \appref{app:alpha_sensitivity}.


\vspace{-4pt}
\subsection{PDI Makes Autonomous Skill Generation Work}
\label{sec:pdi_online_control}

The above analyses treat PDI as a descriptor applied to completed trajectories. A natural question is whether the same signal can improve autonomous skill generation \textit{during} the process itself. To test this, we rerun skill generation, with PDI-guided online intervention explicitly enabled, on the subset of tasks where original trajectories showed persistently low proxy-PDI values. To avoid introducing confounding effects, this subset excludes tasks where the teacher agent succeeded in the first round and those where the PDI monitors remain within healthy thresholds that will not trigger interference.


Concretely, after each reflect step $k$ SPARK computes a weighted proxy-PDI $\hat{d}_k = w_k \cdot \mathrm{PDI}_k$, where $w_k = \min(1,\, k/W)$ is a linear warm-up ramp ($W{=}2$) that suppresses noisy early signals. When $\hat{d}_k < \tau$ ($\tau{=}{-}0.5$), a \emph{soft} intervention injects prompt-only guidance encouraging broader hypothesis revision while keeping the metric hidden from the agent. If two consecutive steps both trigger ($\hat{d}_{k-1} < \tau$ and $\hat{d}_k < \tau$), SPARK escalates to a \emph{strong} intervention that additionally withholds the previous \emph{Next Strategy} section and instructs the agent to anchor on \emph{Verified Facts} and \emph{Current Error Pattern} rather than continuing the stale plan. The full procedure, including the PDI-guided branch, is given in \appref{app:alg_skill}; trajectory-level case studies are also provided in \appref{app:pdi_case_studies}.



\paragraph{Downstream evaluation.}

We evaluate the skills generated \textbf{with and without PDI} control on the same subset of tasks. This ablation directly tests whether the trajectory-level signal can improve skill generation rather than merely a post-hoc descriptor. In~\autoref{tab:pdi_eval}, PDI-guided skills yield stronger pass rates than their non-PDI counterparts. The largest improvement appears on DeepSeek-Chat (pass rate from 9.1\% to 42.4\%). These results indicate that the online intervention consistently improves the downstream utility of distilled skills without introducing negative side effects (case study in~\appref{app:lean4_case_study}). We further compare SPARK against three skill generation baselines (Trace2Skill~\cite{ni2026trace2skill}, AutoRefine~\cite{qiu2026autorefine}, and EvoSkill~\cite{alzubi2026evoskill}) under identical evaluation conditions; SPARK's skill gain exceeds the strongest baseline by $+$0.119 in mean reward (\appref{app:baseline_comparison}).


\subsection{Ablation Study on Task Variants}
\label{sec:ablation}
The main benchmark evaluates one skill on one task instance, leaving robustness to various within-class instances untested. To assess this, we generate 300 task variants via SPARK's pipeline, preserving the overall workflow while purposely resampling task-specific contents. This design tests whether distilled skills encode reusable procedural structure rather than overfitting to instance-specific surface form. In \autoref{tab:variant_transfer}, SPARK retains positive gains that the generated skill improves over the baseline by 3.3--5\% across all stdent models, confirming that distilled skills generalize across instances within the same problem class. For cross-domain generalization, we extend to evaluate SPARK on ALFWorld~\citep{shridhar2021alfworld}, a text-based interactive household environment that differs fundamentally from programming tasks. PDI-refined skills improve overall success rate from 16.7\% to 40.0\%, validating both the SPARK pipeline and PDI in an out-of-distribution setting (\appref{app:alfworld}).

\section{Conclusion}
\label{sec:conclusion}

Skill generation and validation through the lens of trajectory quality are crucial to advance agentic system. In this study, our central contribution is the Posterior Distillation Index (PDI): an interpretable, process-grounded criterion that enables a trajectory-level measure of whether a generated skill is grounded in the task environment rather than relying on prior plans. We demonstrate that the SPARK framework can produce such trajectories and a PDI proxy can guide and improve skill generation online. Together, autonomous skill generation works strongly when agent learns from the environment as a prerequisite, not an afterthought, for producing vital knowledge that future agents can inherit and execute verifiably and reliably.




\bibliographystyle{plainnat}
\bibliography{references}

\clearpage
\appendix

{
\hypersetup{linkcolor=blue!70!black}

\newcommand{\apptocline}[3]{%
  \noindent\hyperref[#3]{\textbf{#1}\hspace{0.6em}#2\,\dotfill\,\pageref{#3}}\par\vspace{1pt}%
}
\newcommand{\apptocsubline}[3]{%
  \noindent\hspace{1.5em}\hyperref[#3]{#1\hspace{0.5em}#2\,\dotfill\,\pageref{#3}}\par\vspace{1pt}%
}

\section*{Appendix Content}
\vspace{0.5em}

\apptocline{A}{Positioning Against Prior Skill-Centric Systems}{app:comparison_table}
\apptocline{B}{Trajectory Feature Correlation Analysis}{app:heatmap}
\apptocline{C}{Case Studies of Online PDI-Guided Control}{app:pdi_case_studies}
\apptocline{D}{External Transfer Case Study: \texttt{lean4-proof}}{app:lean4_case_study}
\apptocline{E}{Sensitivity of PDI to the Smoothing Parameter}{app:alpha_sensitivity}
\apptocline{F}{Sensitivity of PDI to Component Weights}{app:weight_sensitivity}
\apptocline{G}{Exploration Statistics}{app:exploration_stats}
\apptocline{H}{Pipeline Pseudocode}{app:pseudocode}
\apptocsubline{H.1}{Skill Generation Pipeline}{app:alg_skill}
\apptocsubline{H.2}{Task Construction Pipeline}{app:alg_task}
\apptocline{I}{Generated Task Examples}{app:generated_tasks}
\apptocsubline{I.1}{3D Scan Mass Calculation}{app:task_3d}
\apptocsubline{I.2}{WAV Amplitude Analysis}{app:task_wav}
\apptocsubline{I.3}{Weather Station Anomaly Detection}{app:task_weather}
\apptocline{J}{Comparison with Baseline Skill Generation Methods}{app:baseline_comparison}
\apptocline{K}{External Benchmark: ALFWorld}{app:alfworld}
\apptocline{L}{Limitations and Scope}{app:limitations}
}

\clearpage

\section{Positioning Against Prior Skill-Centric Systems}
\label{app:comparison_table}

Table~\ref{tab:comparison} positions SPARK relative to recent skill-centric systems along five dimensions. Prior work has typically focused on only a subset of these properties: some methods emphasize environment-grounded interaction, others study usefulness evaluation or cross-model transfer, but few go beyond demonstrating that skills can be produced or applied. In contrast, SPARK is designed not only to generate transferable skills in executable environments, but also to analyze why some trajectories yield useful downstream skills while others do not. This trajectory-level mechanistic perspective further enables analysis-guided online intervention, making SPARK distinct from prior systems that treat skill generation primarily as a black-box pipeline.

\begin{table*}[h]
  \centering
  \scriptsize
  \setlength{\tabcolsep}{5pt}
  \renewcommand{\arraystretch}{1.15}
  \caption{Comparison with recent skill-centric systems. The highlighted column marks the paper's central focus beyond framework construction: explaining why some trajectories yield useful transferable skills. Binary columns: \textbf{Env.-Grounded Interaction}~: the agent executes in a real environment; \textbf{Usefulness Eval.}~: explicitly evaluates skill-induced $\Delta$performance; \textbf{Cross-Model Transfer}~: skills generated by one model guide a different model; \textbf{Mechanistic Analysis}~: analyzes trajectory-to-skill properties that explain downstream utility; \textbf{Analysis-Guided Intervention}~: uses such analysis to improve skill generation online.}
  \label{tab:comparison}
  \begin{tabular}{@{} l c c c c c @{}}
    \toprule
    \textbf{System} & \textbf{Env.-Grounded} & \textbf{Usefulness} & \textbf{Cross-Model} & \cellcolor{yellow!18}\textbf{Mechanistic} & \textbf{Analysis-Guided} \\
    & \textbf{Interaction} & \textbf{Eval.} & \textbf{Transfer} & \cellcolor{yellow!18}\textbf{Analysis} & \textbf{Intervention} \\
    \midrule
    AgentTrek \citep{xu2025agenttrek}   & \cmark & \xmark & \xmark & \xmark & \xmark \\
    AutoSkill \citep{yang2026autoskill} & \xmark & \xmark & \cmark & \xmark & \xmark \\
    SkillCraft \citep{chen2026skillcraft} & \cmark & \cmark & \xmark & \xmark & \xmark \\
    SkillsBench \citep{li2026skillsbench} & \cmark & \cmark & \cmark & \xmark & \xmark \\
    SkillNet \citep{liang2026skillnet}  & \cmark & \cmark & \cmark & \xmark & \xmark \\
    SkillRL \citep{xia2026skillrl}      & \cmark & \xmark & \xmark & \xmark & \xmark \\
    Trace2Skill \citep{ni2026trace2skill} & \cmark & \cmark & \cmark & \xmark & \xmark \\
    EvoSkill \citep{alzubi2026evoskill} & \cmark & \cmark & \xmark & \xmark & \xmark \\
    AutoRefine \citep{qiu2026autorefine} & \cmark & \cmark & \xmark & \xmark & \xmark \\
    OEL \citep{ye2026oel}               & \cmark & \xmark & \xmark & \xmark & \xmark \\
    SKILL0 \citep{lu2026skill0}         & \cmark & \cmark & \xmark & \xmark & \xmark \\
    \midrule
    \textbf{SPARK (Ours)}               & \cmark & \cmark & \cmark & \cellcolor{yellow!18}\cmark & \cmark \\
    \bottomrule
  \end{tabular}
\end{table*}

\section{Trajectory Feature Correlation Analysis}
\label{app:heatmap}

\begin{figure}[h]
    \centering
    \includegraphics[width=\textwidth]{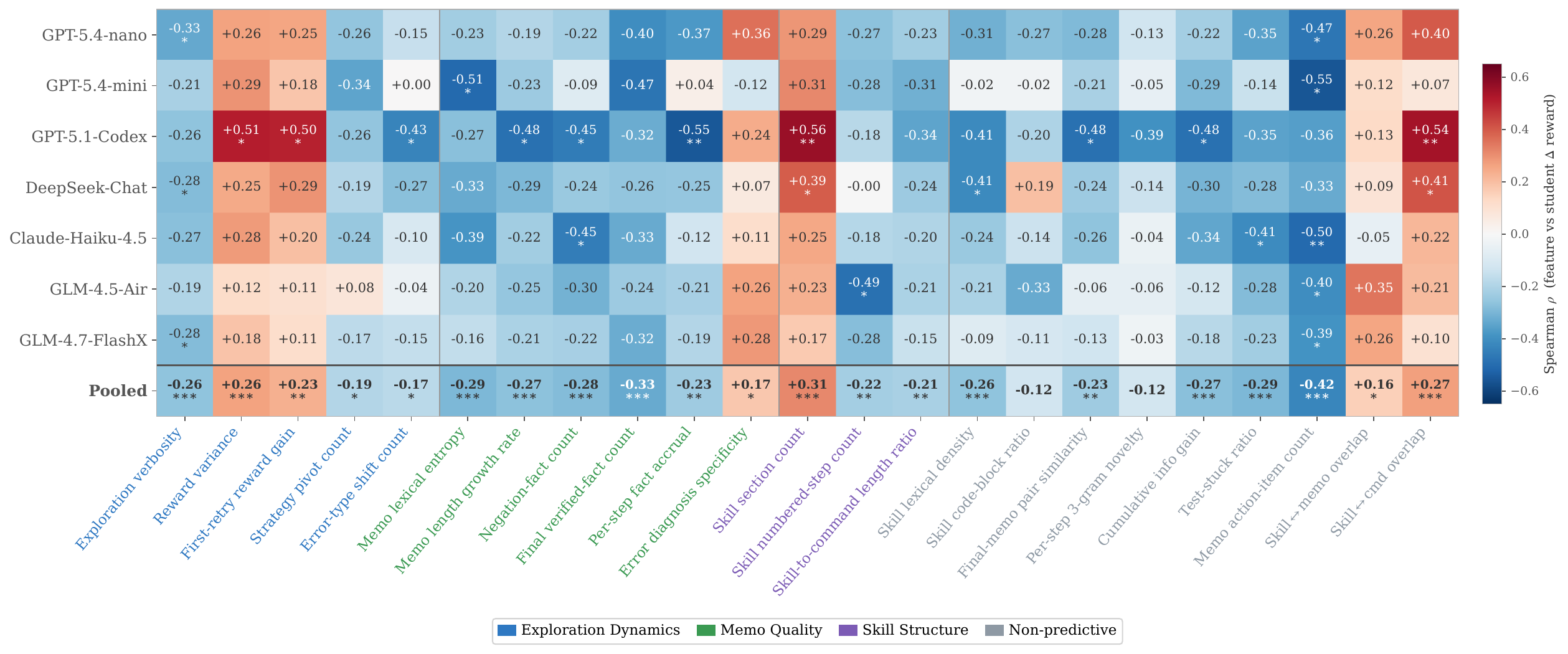}
    \caption{Spearman rank correlation between trajectory-level features (columns) and student skill gain $\Delta r_{m,t}$ (rows). Each row corresponds to a student model; the bottom row pools all (task, model) pairs. Significance: $\ast$ $p<.05$, $\ast\ast$ $p<.01$, $\ast\ast\ast$ $p<.001$.}
    \label{fig:feature_heatmap}
  \end{figure}

  Figure~\ref{fig:feature_heatmap} presents a Spearman rank-correlation heatmap between 23 trajectory-level features and the downstream skill gain $\Delta r_{m,t}$ observed for each student model.
  Each cell reports the Spearman $\rho$ between a feature (column) and $\Delta r_{m,t}$ (row) computed over all tasks where (i)~the baseline student failed ($r^{\text{base}}_{m,t}=0$) and (ii)~a SPARK skill was available.
  The bottom row (\textbf{Pooled}) aggregates all (task, student) pairs across all seven student models.
  Significance levels are annotated as $\ast$ ($p<.05$), $\ast\ast$ ($p<.01$), and $\ast\ast\ast$ ($p<.001$).
  Red cells indicate that higher feature values are associated with larger student gains; blue cells indicate the opposite.

  \paragraph{How to read the figure.}
  Columns are grouped into four categories by colour: \emph{Exploration Dynamics} (blue), \emph{Memo Quality} (green), \emph{Skill Structure} (purple), and \emph{Non-predictive} (grey).
  A feature that shows consistent red across most student rows and a significant pooled $\rho$ is a robust positive predictor of skill usefulness; consistent blue indicates a negative predictor.
  Features in the \emph{Non-predictive} block were hypothesized to matter but showed weak or inconsistent correlations, serving as negative controls.

  \paragraph{Feature definitions.}
  Let $\tau = (\tau_1, \dots, \tau_K)$ denote the sequence of $K$ execution attempts for a task, $\mu_k$ the exploration memo after the $k$-th reflect step, and $s$ the generated skill text.
  We use $\mathrm{tok}(\cdot)$ for word-level tokenization and $\mathrm{J}(a,b) = |W_a \cap W_b| / |W_a \cup W_b|$ for the Jaccard similarity between word sets $W_a$ and $W_b$.

  \subparagraph{Exploration Dynamics.}
  \begin{itemize}[leftmargin=1.2em, itemsep=1pt, topsep=2pt]
    \item \textbf{Compression ratio}: $\rho_c = \sum_k |\tau_k^{\text{stdout}}| \;/\; |s|$. Total raw agent output divided by skill length, matching the main-text definition in \autoref{sec:compression_attempts}.
    \item \textbf{Reward variance}: $\mathrm{Var}(\{r_1, \dots, r_K\})$.
    \item \textbf{First-retry reward gain}: $\Delta r^{(1)} = r_2 - r_1$.
    \item \textbf{Strategy pivot count}: $\sum_{k=2}^{K} \mathbf{1}[\mathrm{J}(\text{strategy}_{k-1}, \text{strategy}_k) < 0.15]$, where $\text{strategy}_k$ is the ``Next Strategy'' section of $\mu_k$.
    \item \textbf{Error-type shift count}: same as above, applied to the ``Current Error Pattern'' section.
  \end{itemize}

  \subparagraph{Memo Quality.}
  \begin{itemize}[leftmargin=1.2em, itemsep=1pt, topsep=2pt]
    \item \textbf{Memo lexical entropy}: $\bar{H} = \frac{1}{K}\sum_k H(\mu_k)$, where $H(\mu_k) = -\sum_w p_w \log_2 p_w$ is the Shannon entropy over word frequencies.
    \item \textbf{Memo length growth rate}: $(|\mu_K| - |\mu_1|) / |\mu_1|$. Fractional increase in memo character count.
    \item \textbf{Negation-fact count}: number of bullet items in the final ``Verified Facts'' section matching negation keywords (\texttt{not}, \texttt{never}, \texttt{failed}, \texttt{wrong}, etc.).
    \item \textbf{Final verified-fact count}: total bullet items in the final ``Verified Facts'' section.
    \item \textbf{Per-step fact accrual}: $\frac{1}{K{-}1}\sum_{k=2}^{K}(n_k^{\text{facts}} - n_{k-1}^{\text{facts}})$.
    \item \textbf{Error diagnosis specificity}: mean ratio of concrete references (inline code, numbers, file paths) to total words in the ``Current Error Pattern'' section across all memos.
  \end{itemize}

  \subparagraph{Skill Structure.}
  \begin{itemize}[leftmargin=1.2em, itemsep=1pt, topsep=2pt]
    \item \textbf{Skill section count}: number of Markdown heading lines (\texttt{\#}, \texttt{\#\#}, \dots) in $s$.
    \item \textbf{Skill numbered-step count}: number of lines matching the pattern \texttt{/\textasciicircum\textbackslash d+[.)]/} in $s$.
    \item \textbf{Skill-to-command length ratio}: $|s| \;/\; \overline{|\mathbf{c}_{\text{succ}}|}$, where $\overline{|\mathbf{c}_{\text{succ}}|}$ is the mean character length of successful-attempt agent commands.
  \end{itemize}

  \subparagraph{Non-predictive (negative controls).}
  \begin{itemize}[leftmargin=1.2em, itemsep=1pt, topsep=2pt]
    \item \textbf{Skill lexical density}: $|\mathrm{unique}(\mathrm{tok}(s))| / |\mathrm{tok}(s)|$.
    \item \textbf{Skill code-block ratio}: total characters inside \texttt{```\dots```} blocks divided by $|s|$.
    \item \textbf{Final-memo pair similarity}: $\mathrm{J}(\mu_{K-1}, \mu_K)$.
    \item \textbf{Per-step 3-gram novelty}: mean fraction of new word 3-grams at each memo step.
    \item \textbf{Cumulative info gain}: sum of per-step 3-gram novelty across all steps.
    \item \textbf{Test-stuck ratio}: fraction of consecutive attempt pairs with identical test summaries.
    \item \textbf{Memo action-item count}: mean number of bullet/numbered items per memo.
    \item \textbf{Skill$\leftrightarrow$memo overlap}: $\mathrm{J}(s, \mu_K)$.
    \item \textbf{Skill$\leftrightarrow$cmd overlap}: $\mathrm{J}(s, \mathbf{c}_{\text{succ}})$.
  \end{itemize}

\section{Case Studies of Online PDI-Guided Control}
\label{app:pdi_case_studies}

We provide two trajectory-level case studies that compare PDI-enabled runs against observe-only controls on \texttt{3d-scan-calc} and \texttt{manufacturing-codebook-normalization}. In \texttt{3d-scan-calc}, the weighted proxy-PDI triggers a soft intervention at step~4 and escalates to a strong one at step~5; the trajectory then rebounds to a clearly positive regime, whereas the observe-only control keeps drifting downward. In \texttt{manufacturing-codebook-normalization}, two earlier soft triggers (steps~3 and~5) already suffice to reverse the decline and sustain positive proxy-PDI, while the control again stays persistently negative. Together, these cases show that proxy-PDI is not merely descriptive but an actionable online signal: it detects plan-repetitive or stale-summary exploration and supports graded, prompt-only interventions that redirect search without exposing the metric to the model.

\begin{figure*}[h]
  \centering
  \includegraphics[width=\textwidth]{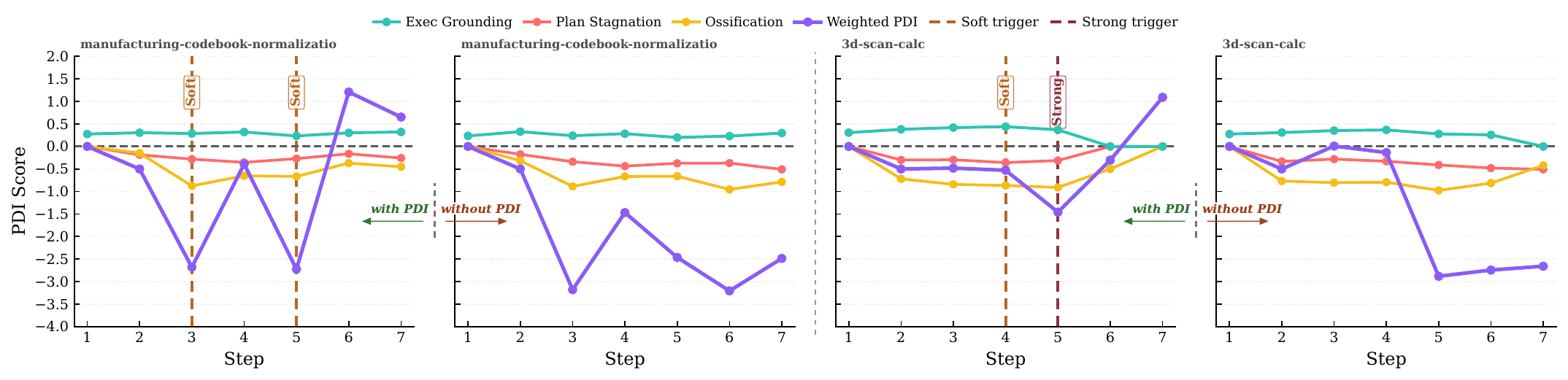}
  \caption{Two case studies of online PDI-guided control, comparing PDI-enabled runs against observe-only controls on \texttt{3d-scan-calc} and \texttt{manufacturing-codebook-normalization}. Each panel plots execution grounding ($\phi_{\mathrm{exec}}$), plan copying ($\phi_{\mathrm{plan}}$), memo ossification ($\phi_{\mathrm{oss}}$), and the warmup-weighted proxy-PDI used for intervention decisions. Vertical dashed lines mark soft and strong triggers.}
  \label{fig:pdi_case_studies}
\end{figure*}

\section{External Transfer Case Study: \texttt{lean4-proof}}
\label{app:lean4_case_study}

Among all tasks evaluated under PDI-refined skills, \texttt{lean4-proof} exhibits the clearest behavioural shift in a held-out student model (Claude Haiku~4.5, never used during skill generation or PDI refinement).
The task requires proving $S\,n \le 2$ for the geometric-series sequence $S\,0 = 1,\; S\,(n{+}1) = S\,n + 1/2^{n+1}$ in Lean~4; the proof body must start at line~15 of \texttt{solution.lean}, and the verifier runs \texttt{lake env lean -DwarningAsError=true solution.lean} with zero output indicating success.
We present the analysis in three tables that progressively zoom in: skill-level differences (\autoref{tab:lean4_skill_diff}), step-by-step trajectory comparison (\autoref{tab:lean4_trajectory}), and the resulting strategy and score changes (\autoref{tab:lean4_strategy_score}).

\begin{table}[h]
  \centering
  \scriptsize
  \setlength{\tabcolsep}{3pt}
  \renewcommand{\arraystretch}{1.25}
  \caption{Structural comparison between the original and PDI-refined skill documents for \texttt{lean4-proof}. \textcolor{red}{Red text} highlights the components that most directly alter the student's execution strategy. The PDI-refined skill is qualitatively different: it encodes a \emph{complete reference implementation} with the exact \texttt{norm\_num}\,/\,\texttt{ring}\,/\,\texttt{linarith} tactic chain, rather than a high-level proof sketch.}
  \label{tab:lean4_skill_diff}
  \begin{tabular}{@{} p{2.6cm} >{\raggedright\arraybackslash}p{5.2cm} >{\raggedright\arraybackslash}p{5.8cm} @{}}
    \toprule
    \textbf{Dimension} & \textbf{Original Skill (non-PDI)} & \textbf{PDI-Refined Skill} \\
    \midrule
    Document length
      & 118 lines
      & \textbf{393 lines} ($3.3\times$) \\[3pt]
    Structural modules
      & 4 (overview, orientation, strategy, pitfalls)
      & \textbf{7} (+ step-by-step workflow, reference proof, testing) \\[3pt]
    Contains executable proof?
      & No
      & \textcolor{red}{\textbf{Yes}, complete \texttt{theorem} block with exact tactic sequence} \\[3pt]
    Verification command
      & Partial (\texttt{lake env lean solution.lean})
      & \textcolor{red}{\textbf{Exact}: \texttt{lake env lean \mbox{-DwarningAsError=true} solution.lean}} \\[3pt]
    ``Write to \texttt{/tmp} first'' workflow
      & Not mentioned
      & \textcolor{red}{\textbf{Yes}, Step~3 of workflow prevents corrupting the solution file} \\
    \bottomrule
  \end{tabular}
\end{table}

\begin{table*}[h]
  \centering
  \scriptsize
  \setlength{\tabcolsep}{3pt}
  \renewcommand{\arraystretch}{1.22}
  \caption{Step-by-step trajectory comparison for \texttt{lean4-proof} on Claude Haiku~4.5. \textcolor{red}{Red text} marks the critical behavioural differences induced by the PDI-refined skill. The non-PDI agent spends 15 commands exploring the environment and iterating on broken proofs; the PDI agent follows the skill's recipe in 5 commands with zero failures.}
  \label{tab:lean4_trajectory}
  \begin{tabular}{@{} c >{\raggedright\arraybackslash}p{6.4cm} >{\raggedright\arraybackslash}p{6.4cm} @{}}
    \toprule
    \textbf{Step} & \textbf{non-PDI Trajectory} (15 cmds, 3 failures) & \textbf{PDI Trajectory} (5 cmds, 0 failures) \\
    \midrule
    0
      & \texttt{cat /root/spark-skills/*/SKILL.md} \newline reads skill (high-level sketch only)
      & \texttt{cat /root/spark-skills/*/SKILL.md} \newline reads PDI skill (\textcolor{red}{contains full proof}) \\[4pt]
    1
      & \texttt{cat -n solution.lean} \newline reads template
      & \texttt{cat -n solution.lean} \newline reads template \\[4pt]
    2
      & \texttt{cat lean-toolchain} \newline checks toolchain version
      & \texttt{cat -n /app/baseline/solution.lean} \newline \textcolor{red}{verifies immutable prefix} \\[4pt]
    3
      & \texttt{find Library -name "*.lean"} \newline explores library structure
      & Writes proof to \texttt{/tmp/test.lean}, then copies to \texttt{solution.lean} \newline \textcolor{red}{follows ``test in /tmp first'' recipe} \\[4pt]
    4
      & Writes proof attempt \#1 $\to$ \textbf{FAIL} (type errors)
      & \texttt{lake env lean \mbox{-DwarningAsError=true} solution.lean} \newline \textcolor{red}{\textbf{SUCCESS} (exit 0)} \\[4pt]
    5--14
      & Iterates through attempts \#2--\#6, each failing with tactic or type errors; 3 total failures
      & \\
    \bottomrule
  \end{tabular}
\end{table*}

\begin{table*}[h]
  \centering
  \renewcommand{\arraystretch}{1.22}
  \caption{Strategy shift and quantitative changes induced by the PDI-refined skill on \texttt{lean4-proof}. \textcolor{red}{Red text} marks the PDI-specific cues that drive each behavioural change. The bottom rows report cross-model PDI results: the largest gains appear on weaker models (GPT-5.1-Codex, GLM-4.5-Air), consistent with the hypothesis that PDI-refined skills compensate for capability gaps by converting posterior evidence into executable recipes.}
  \label{tab:lean4_strategy_score}
  \resizebox{\textwidth}{!}{%
  \scriptsize
  \begin{tabular}{@{} p{1.5cm} >{\raggedright\arraybackslash}p{3.8cm} >{\raggedright\arraybackslash}p{4.0cm} >{\raggedright\arraybackslash}p{2.8cm} p{1.5cm} @{}}
    \toprule
    \textbf{Phase} & \textbf{Original Skill Cue} & \textbf{PDI Skill Cue} & \textbf{Strategy Shift} & \textbf{Metric} \\
    \midrule
    Task framing
      & High-level proof sketch: derive closed form, then prove bound
      & \textcolor{red}{Complete tactic recipe} with exact \texttt{simp~[S]} + \texttt{ring} pattern
      & From re-deriving the plan to direct theorem construction
      & \\[4pt]
    Proof construction
      & Agent guesses tactic sequences; no reference implementation
      & \textcolor{red}{Reference proof block} with \texttt{norm\_num} / \texttt{ring} / \texttt{linarith} chain
      & From trial-and-error to copy-and-adapt
      & Failed cmds: $3 \!\to\! \mathbf{0}$ \\[4pt]
    Verification
      & \texttt{lake env lean solution.lean} (missing \texttt{-DwarningAsError})
      & \textcolor{red}{Exact command}: \texttt{lake env lean \mbox{-DwarningAsError=true}}
      & Correct flags on first attempt
      & Total cmds: $15 \!\to\! \mathbf{5}$ \\[4pt]
    Safety workflow
      & Writes directly to \texttt{solution.lean}
      & \textcolor{red}{Writes to \texttt{/tmp} first}, copies after verification
      & Prevents corrupting the immutable prefix
      & Msgs: $16 \!\to\! \mathbf{6}$ \\
    \midrule
    \multicolumn{3}{@{}l}{\textbf{Cross-model PDI results on \texttt{lean4-proof}}} & \multicolumn{2}{l}{} \\
    \midrule
    \multicolumn{3}{@{}l}{GPT-5.1-Codex: non-PDI FAIL $\to$ PDI \textbf{PASS}} & \multicolumn{2}{l}{$\Delta = +1.0$} \\
    \multicolumn{3}{@{}l}{GLM-4.5-Air: non-PDI FAIL $\to$ PDI \textbf{PASS}} & \multicolumn{2}{l}{$\Delta = +1.0$} \\
    \multicolumn{3}{@{}l}{Claude Haiku~4.5: stochastic (FAIL/PASS across runs)} & \multicolumn{2}{l}{$\Delta \in \{0, +1.0\}$} \\
    \bottomrule
  \end{tabular}%
  }
\end{table*}

\paragraph{Why PDI made the difference.}
The non-PDI skill tells the agent \emph{what} to prove (closed-form identity $\to$ bound) but not \emph{how} to write the Lean~4 tactics.
The PDI-refined skill was generated after observing that weaker models fail precisely at the tactic level, so it includes:
(1)~the exact \texttt{simp [S]} + \texttt{ring} pattern for the inductive step;
(2)~the \texttt{norm\_num} + \texttt{linarith} pattern for the bound derivation; and
(3)~a ``test in \texttt{/tmp} first'' workflow that prevents corrupting the solution file.
This is the PDI mechanism in action: by comparing successful and failed trajectories across models, the refinement process identifies \emph{where agents actually get stuck} and injects targeted procedural knowledge.
The result is not merely a longer document, but a fundamentally different \emph{action policy} for the student: the original skill transfers a plan; the PDI-refined skill transfers an execution recipe anchored in verified evidence.

\section{Sensitivity of PDI to the Smoothing Parameter}
\label{app:alpha_sensitivity}

\begin{figure}[h]
  \centering
  \includegraphics[width=0.95\linewidth]{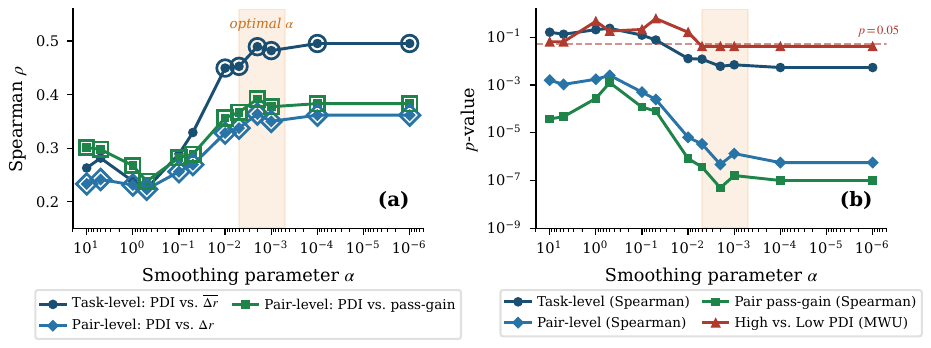}
  \caption{Sensitivity of PDI to the smoothing parameter $\alpha$. (a)~Spearman $\rho$ between PDI and three outcome measures; circled points are significant at $p{<}0.05$. (b)~Corresponding $p$-values on a log scale; the red dashed line marks $p{=}0.05$. The shaded band highlights the optimal region $\alpha \in [5{\times}10^{-4},\, 5{\times}10^{-3}]$.}
  \label{fig:alpha_sensitivity}
\end{figure}

The smoothing constant $\alpha$ in \autoref{sec:prior_posterior} controls the sharpness of the empirical token distributions: smaller $\alpha$ yields distributions closer to raw frequency counts, while larger $\alpha$ flattens them toward uniform. \autoref{fig:alpha_sensitivity} reports the Spearman $\rho$ and $p$-values of PDI across $\alpha \in [10^{-10}, 10]$. All three correlation metrics (task-level, pair-level, and pass-gain) exhibit a plateau for $\alpha \lesssim 10^{-3}$, where smoothing is negligible relative to observed counts, and degrade monotonically as $\alpha$ increases beyond $10^{-2}$. The Mann--Whitney U test between high- and low-PDI groups remains significant ($p{<}0.05$) for $\alpha \leq 0.005$ and loses significance for $\alpha \geq 0.007$. This pattern confirms that PDI is robust to the choice of $\alpha$ within a broad low-smoothing regime.

\section{Sensitivity of PDI to Component Weights}
\label{app:weight_sensitivity}

PDI combines three z-scored components with equal weights: $\mathrm{PDI} = z(\phi_{\mathrm{exec}}) - z(\phi_{\mathrm{plan}}) - z(\phi_{\mathrm{oss}})$. We examine whether this choice is justified by testing (i)~whether the directional structure itself is robust, (ii)~whether fitted weights generalize across held-out models and tasks, and (iii)~whether each component is necessary.

\paragraph{Directional structure.}
We sweep all weight combinations $(w_e, w_p, w_o)$ with $w_e > 0$ (preserving the sign convention that execution grounding contributes positively while plan copying and memo ossification contribute negatively). The large majority of sign-consistent combinations yield statistically significant Spearman correlation ($p{<}0.05$) with student skill gain, indicating that PDI's predictive power derives from its directional structure rather than from any particular weight assignment.

\paragraph{Cross-validated weight stability.}
To test whether fitted weights generalize, we perform 5-fold task cross-validation: on each fold, we grid-search the weight vector that maximizes Spearman $\rho$ on the training tasks and evaluate it on the held-out fold. \autoref{tab:weight_cv} reports the results. Grid search produces four distinct ``optimal'' weight vectors across five folds, yet equal weighting matches or exceeds the fitted optimum on every fold (5/5). A complementary leave-one-model-out protocol yields a similar pattern: nine folds produce five distinct optimal vectors. Together, these results confirm that weight tuning does not yield transferable gains and that the fitted optimum is an artifact of the particular train split.

\begin{table}[h]
  \centering
  \small
  \setlength{\tabcolsep}{5pt}
  \renewcommand{\arraystretch}{1.15}
  \caption{5-fold task cross-validation of PDI component weights. On each fold, the weight vector maximizing $\rho$ on training tasks is evaluated on held-out tasks. Equal weighting ($1,1,1$) matches or outperforms the fitted optimum on every fold, while the fitted vectors are unstable across folds.}
  \label{tab:weight_cv}
  \begin{tabular}{@{} c l c c c @{}}
    \toprule
    \textbf{Fold} & \textbf{Fitted $(w_e,w_p,w_o)$} & \textbf{$\rho_{\text{fitted}}$} & \textbf{$\rho_{\text{equal}}$} & \textbf{Winner} \\
    \midrule
    1 & $(1.5,\; 0.5,\; 1.0)$ & 0.153 & \textbf{0.241} & Equal \\
    2 & $(1.0,\; 0.5,\; 0.5)$ & 0.179 & \textbf{0.179} & Tie \\
    3 & $(1.0,\; 0.0,\; 0.5)$ & 0.426 & \textbf{0.492} & Equal \\
    4 & $(0.5,\; 0.0,\; 1.0)$ & 0.274 & \textbf{0.281} & Equal \\
    5 & $(0.5,\; 0.0,\; 0.5)$ & 0.252 & \textbf{0.252} & Tie \\
    \midrule
    \textbf{Mean} & \multicolumn{1}{c}{4 distinct vectors} & 0.257 & \textbf{0.289} & \textbf{Equal} \\
    \bottomrule
  \end{tabular}
\end{table}

\paragraph{Component ablation.}
Removing execution grounding ($w_e{=}0$) collapses held-out correlation to near zero; removing memo ossification ($w_o{=}0$) also degrades predictive power. Equal weighting thus serves as a principled, parameter-free default that preserves all three empirically validated directional signals without introducing tunable hyperparameters that risk overfitting to a particular task or model distribution.

\section{Exploration Statistics}
\label{app:exploration_stats}

Table~\ref{tab:exploration_stats} summarises the exploration cost of the SPARK skill-generation pipeline across four teacher models. All models share the same task pool and Docker-based execution environment; the maximum retry budget $N_{\max}$ is 7 for Claude Opus 4.6, GPT-5.2-Codex, GPT-5.4 and GLM-5. These figures should be interpreted as an upfront investment in capability transfer rather than a per-query serving cost: strong interactive teachers are expensive to run repeatedly, but once their trajectories are distilled into reusable \texttt{SKILL.md} documents, the same procedural knowledge can be deployed to much cheaper student models at inference time. In this sense, SPARK trades one-time high-cost exploration for repeated low-cost execution. The main-text results show that this trade can be favorable not only economically but also in effectiveness, since several skill-equipped students even surpass the interaction-free performance of the stronger teacher models.

\begin{table}[h]
  \centering
  \small
  \setlength{\tabcolsep}{4pt}
  \renewcommand{\arraystretch}{1.15}
  \caption{Exploration statistics of the SPARK skill-generation pipeline. Agent execution time excludes Docker build and teardown overhead. Interaction turns count agent messages plus shell commands per attempt. Cost estimates for GLM-5 and GPT-5.2-Codex are calibrated from execution time.}
  \label{tab:exploration_stats}
  \begin{tabular}{@{} l c c c c @{}}
    \toprule
    \textbf{Metric}
      & \textbf{Claude Opus 4.6}
      & \textbf{GPT-5.4}
      & \textbf{GLM-5}
      & \textbf{GPT-5.2-Codex} \\
    \midrule
    Avg.\ attempts / task        & 6.00  & 3.93  & 3.25  & 5.37  \\
    Avg.\ reward (all attempts)  & 0.120 & 0.246 & 0.394 & 0.080 \\
    Avg.\ agent exec.\ time      & 8m50s & 7m04s & 4m20s & 6m13s \\
    Avg.\ interaction turns      & 30.5  & 32.3  & 16.2 & 23.2 \\
    Avg.\ cost / attempt (\$)    & 4.11  & 5.70  & 3.03 & 2.70 \\
    Avg.\ cost / task (\$)       & 24.72 & 22.35 & 9.90 & 14.49 \\
    \bottomrule
  \end{tabular}
\end{table}

\paragraph{Student inference cost and amortization.}
While teacher exploration is expensive, the distilled skills are consumed by much cheaper student models at inference time. \autoref{tab:student_cost} reports the per-task inference cost of representative student models, computed from the token-level usage logs recorded in each evaluation trajectory. Across all measured students, the per-task cost ranges from \$0.02 (DeepSeek-Chat) to \$0.18 (GLM-4.5-Air), yielding 86-task totals between \$1.89 and \$15.48. Compared with the cheapest teacher (GLM-5 at \$9.90/task), even the most expensive student is $55\times$ cheaper per task; DeepSeek-Chat is over $1{,}100\times$ cheaper than Claude Opus~4.6. A single teacher exploration run (\$852--\$2{,}126 for 86 tasks) can therefore be amortized over hundreds of student deployments at negligible marginal cost.

\begin{table}[h]
  \centering
  \small
  \setlength{\tabcolsep}{4pt}
  \renewcommand{\arraystretch}{1.15}
  \caption{Student model inference cost per task, computed from token-level usage logs in evaluation trajectories. The teacher/student ratio uses the cheapest teacher (Claude Opus~4.6 at \$24.72/task) as the numerator.}
  \label{tab:student_cost}
  \begin{tabular}{@{} l c c c @{}}
    \toprule
    \textbf{Student Model} & \textbf{Avg.\ cost / task (\$)} & \textbf{86-task total (\$)} & \textbf{Teacher / Student ratio} \\
    \midrule
    DeepSeek-Chat   & 0.022 & 1.89  & $1{,}124\times$ \\
    GLM-4.7-FlashX  & 0.085 & 7.31  & $291\times$ \\
    GLM-4.5-Air     & 0.180 & 15.48 & $137\times$ \\
    \bottomrule
  \end{tabular}
\end{table}

\section{Pipeline Pseudocode}
\label{app:pseudocode}

  \subsection{Skill Generation Pipeline}
  \label{app:alg_skill}

  Algorithm~\ref{alg:skill_gen} describes the iterative skill-generation loop introduced in \autoref{sec:skill_gen}. The teacher model explores each task inside a containerized sandbox for up to $N_{\max}$ attempts. Failed attempts trigger a \textsc{Reflect} step that rewrites the exploration memo; a successful attempt triggers evidence assembly and skill distillation.

  \begin{algorithm}[h]
  \caption{SPARK Skill Generation Pipeline}
  \label{alg:skill_gen}
  \begin{algorithmic}[1]
  \Require Task set $\mathcal{T}$, teacher model $\mathcal{M}_{\mathrm{teach}}$, max attempts $N_{\max}$, parallelism $P$, token budgets $\mathcal{B}$
  \Ensure Skill repository $\mathcal{S} = \{s_t\}_{t \in \mathcal{T}_{\mathrm{solved}}}$

  \State Clean stale Docker artifacts; prefetch shared base images
  \State $\mathcal{T}_{\mathrm{active}} \gets \Call{ResolveTasks}{\mathcal{T}}$ \Comment{filter blacklist, skip completed}

  \For{each task $t \in \mathcal{T}_{\mathrm{active}}$ \textbf{in parallel} ($P$ workers)}
    \State Initialise exploration memo $\mu_0 \gets \varnothing$; attempt counter $k \gets 0$
    \While{$k < N_{\max}$}
      \State $k \gets k + 1$
      \State \textbf{// Stage 1: Execute}
      \State Prepare staging directory: copy task files, inject $\mu_{k-1}$ into \texttt{instruction.md}
      \If{$k = N_{\max}$}
        \State Append urgency signal to injection \Comment{final-attempt warning}
      \EndIf
      \State $\tau_k \gets \Call{Execute}{t, \mathcal{M}_{\mathrm{teach}}}$ via Docker sandbox
      \State \textbf{// Stage 2: Judge}
      \State $(r_k, \mathbf{c}_k, \sigma_k) \gets \Call{Judge}{\tau_k}$ \Comment{reward, commands, test summary}
      \If{$r_k \geq 1.0$} \Comment{task solved}
        \State \textbf{// Stage 3: Distill}
        \State $\mathcal{E} \gets \Call{AssembleEvidence}{t, \tau_{1:k}, \mu_{0:k-1}}$ \Comment{six evidence blocks}
        \State $s_t \gets \Call{GenerateSkill}{\mathcal{E}, \mathcal{B}}$ via $\mathcal{M}_{\mathrm{teach}}$
        \State Save $s_t$ as \texttt{SKILL.md}; record trajectory; \textbf{break}
      \Else
        \State \textbf{// Stage 3: Reflect}
        \State $\mu_k \gets \Call{Reflect}{\mu_{k-1}, \mathbf{c}_k, \sigma_k}$ via $\mathcal{M}_{\mathrm{teach}}$ \Comment{rewrite, not append}
        \State \textbf{// Stage 4: PDI-guided intervention}
        \State $w_k \gets \min(1,\; k / W)$ \Comment{linear warm-up, $W{=}2$}
        \State $\hat{d}_k \gets w_k \cdot \mathrm{PDI}_k(\mu_k, \mu_{k-1}, \mathbf{c}_k, \sigma_k, \sigma_{k-1})$
        \If{$\hat{d}_k < \tau$} \Comment{$\tau{=}{-}0.5$}
          \If{$k > 1$ \textbf{and} $\hat{d}_{k-1} < \tau$}
            \State Inject \textsc{StrongIntervention} into next retry prompt
          \Else
            \State Inject \textsc{SoftIntervention} into next retry prompt
          \EndIf
        \EndIf
      \EndIf
    \EndWhile
    \If{task unsolved after $N_{\max}$ attempts}
      \State Save error report with final memo $\mu_{N_{\max}}$
    \EndIf
  \EndFor

  \State \textbf{// Evaluation}
  \For{each student model $\mathcal{M}_s \in \{\mathcal{M}_1, \dots, \mathcal{M}_S\}$}
    \State Run baseline (no skill), SPARK skill, and human skill conditions
    \State Compute per-task $\Delta r_{s,t}$ and aggregate statistics
  \EndFor
  \end{algorithmic}
  \end{algorithm}

  \subsection{Task Construction Pipeline}
  \label{app:alg_task}

  Algorithm~\ref{alg:task_gen} details the task-construction pipeline (\autoref{sec:task_construction}). A natural-language prompt is expanded into a structured blueprint, iteratively critiqued and repaired, rendered into a runnable task directory, and validated by executing the oracle inside the target Docker environment.

  \begin{algorithm}[h]
  \caption{SPARK Task Construction Pipeline}
  \label{alg:task_gen}
  \begin{algorithmic}[1]
  \Require Prompt specification $\mathcal{P}$ (text, tool hints, constraints), LLM $\mathcal{M}$, tool catalogue $\mathcal{C}$, max revisions $R_{\max}$, schema retries $S_{\max}$
  \Ensure Validated task directory $\mathcal{D}_t$ or failure report

  \State \textbf{// Phase 1: Blueprint Generation}
  \State Compose generation prompt from $\mathcal{P}$, $\mathcal{C}$, and Harbor format specification
  \For{$i = 1$ \textbf{to} $S_{\max}$}
    \State $b \gets \Call{CallLLM}{\mathcal{M}, \text{prompt}}$; parse JSON $\to$ \textsc{TaskBlueprint}
    \If{schema validation passes}
      \State \textbf{break}
    \EndIf
  \EndFor

  \State \textbf{// Phase 2: Iterative Critique--Repair--Validate}
  \State $\mathit{critique\_repairs} \gets 0$
  \For{$j = 0$ \textbf{to} $R_{\max}$}
    \State \textbf{// Critique}
    \State $I_{\mathrm{det}} \gets \Call{DeterministicChecks}{b, \mathcal{P}}$ \Comment{output-path consistency, evidence validity}
    \State $I_{\mathrm{llm}} \gets \Call{LLMCritique}{b, \mathcal{P}}$ \Comment{answer leakage, oracle--verifier alignment}
    \State $I \gets I_{\mathrm{det}} \cup I_{\mathrm{llm}}$
    \If{$I$ contains blocking issues \textbf{and} $\mathit{critique\_repairs} < 1$}
      \State $b \gets \Call{Repair}{b, I}$; $\mathit{critique\_repairs} \gets \mathit{critique\_repairs} + 1$; \textbf{continue}
    \EndIf
    \State \textbf{// Render}
    \State $\mathcal{D}_t \gets \Call{Render}{b}$ \Comment{instruction.md, Dockerfile, oracle, verifier, support files}
    \State \textbf{// Validate}
    \State Run \texttt{harbor tasks check} on $\mathcal{D}_t$ \Comment{static structure check}
    \State $v \gets \Call{RunOracle}{\mathcal{D}_t}$ \Comment{execute oracle in Docker, run pytest verifier}
    \If{$v.\mathit{passed}$ \textbf{and} no blocking issues in $I$}
      \State Record generation trace; \Return $\mathcal{D}_t$ \Comment{success}
    \EndIf
    \If{$j < R_{\max}$}
      \State $b \gets \Call{Repair}{b, I \cup v.\mathit{feedback}}$ \Comment{repair from critique + validation}
    \EndIf
  \EndFor
  \State Record trace with all attempts; \Return failure report
  \end{algorithmic}
  \end{algorithm}

\subsubsection{Integrated Techniques for Exploration Quality}
\label{app:techniques}

SPARK integrates several complementary techniques that improve exploration efficiency and distillation fidelity. To reduce noise in the skill-generation context, it applies a lightweight command-processing pipeline: each command is (i)~classified into one of five semantic categories (\textsc{Verify}, \textsc{Implement}, \textsc{Inspect}, \textsc{Prepare}, \textsc{Action}), (ii)~scored by a keyword-weighted importance function that prioritizes verification and implementation actions, and (iii)~filtered to remove low-signal commands (e.g., \texttt{ls}, \texttt{pwd}, \texttt{echo}). Multi-line commands that match known domain patterns (e.g., spreadsheet manipulation via \texttt{openpyxl}, PDDL planning, bibliographic API queries) are replaced with semantic summaries. The top-$k$ commands (default $k{=}12$) are selected by score and presented in execution order, yielding a concise yet high-signal execution chain. In addition, as the agent approaches $N_{\max}$, retry injection adds an escalating urgency signal.

\section{Generated Task Examples}
\label{app:generated_tasks}

We present three tasks produced by the SPARK task-construction pipeline (\autoref{sec:task_construction}) to illustrate the diversity and self-contained nature of the generated benchmarks. Each task was created from a short natural-language prompt, automatically expanded into a full blueprint, critiqued, repaired, and validated by executing the oracle inside the target Docker environment. No manual editing was performed after generation.

  \subsection{3D Scan Mass Calculation}
  \label{app:task_3d}

  \paragraph{Prompt (input).} \emph{``Calculate the mass of a 3D printed part from a binary STL scan. The 2-byte Attribute Byte Count field stores the Material ID. Filter out scanning debris by keeping only the largest connected component, look up the density, and output the mass.''}

  \paragraph{Generated task.} The pipeline produces a complete task directory containing:
  (i)~an \texttt{instruction.md} that asks the agent to parse a binary STL file, identify the largest connected component of the triangle mesh, extract the Material ID from the attribute field, look up the corresponding density from a provided material table, compute the volume via the signed-tetrahedron method, and write the result to a JSON file;
  (ii)~a \texttt{build\_data.py} script that deterministically synthesizes the STL asset and material table during Docker build;
  (iii)~a pytest oracle that checks the reported mass within 0.1\% relative tolerance.

  This task exercises binary file parsing, computational geometry (connected components, mesh volume), and multi-source data integration, a combination unlikely to be solved by template matching alone.

  \subsection{WAV Amplitude Analysis}
  \label{app:task_wav}

  \paragraph{Prompt (input).} \emph{``Parse a binary WAV file, split it into 1-second segments, compute the RMS amplitude of each segment, and identify the loudest one.''}

  \paragraph{Generated task.} The agent must read a WAV file using only Python's standard library, segment the audio by sample rate, compute per-segment RMS values, and produce a structured JSON report including sample rate, channel count, duration, per-segment statistics, and the index of the loudest segment. The oracle verifies all numeric fields within 0.01\% relative tolerance. The audio asset is synthesized deterministically during environment build, ensuring reproducibility without distributing binary data.

  \subsection{Weather Station Anomaly Detection}
  \label{app:task_weather}

  \paragraph{Prompt (input).} \emph{``Detect anomalous hourly temperature readings across multiple weather stations using a 3$\sigma$ threshold.''}

  \paragraph{Generated task.} The pipeline generates a CSV dataset containing hourly readings from multiple stations with injected anomalies. The agent must compute per-station population statistics, flag readings beyond three standard deviations, and output a JSON report with station count, total readings, anomaly count, and a timestamp-sorted anomaly list. The oracle validates both aggregate counts and per-anomaly fields. This task requires careful attention to specification details (population vs.\ sample standard deviation, strict vs.\ non-strict inequality, and per-station independence) that are common sources of silent errors.

  \paragraph{Observations.} Across all three examples, the pipeline consistently produces tasks that are (i)~fully self-contained (all assets are built deterministically at Docker build time), (ii)~oracle-validated (the solution script passes the pytest verifier before the task is accepted), and (iii)~non-trivial (each requires domain-specific reasoning beyond simple text manipulation). These properties make the generated tasks suitable for both skill-generation training and downstream evaluation.

\section{Comparison with Baseline Skill Generation Methods}
\label{app:baseline_comparison}

To further validate the effectiveness of PDI-guided SPARK, we compare against three recent skill generation methods, Trace2Skill~\citep{ni2026trace2skill}, AutoRefine~\citep{qiu2026autorefine}, and EvoSkill~\citep{alzubi2026evoskill}, on the SkillsBench task suite. All methods use the same skill generation model (\texttt{claude-sonnet-4}) and the same evaluation pipeline (student model: \texttt{deepseek-chat}; agent framework: \texttt{qwen-coder}; timeout multiplier: 0.5; Docker-based Harbor sandbox). The skill gain $\Delta$ is the primary comparison metric, as it controls for baseline stochasticity across independent runs.

\paragraph{Reproduction details.}
We re-implement each baseline following its published algorithm. Key configuration parameters are summarized in \autoref{tab:baseline_config}.

\begin{table}[h]
  \centering
  \small
  \setlength{\tabcolsep}{4pt}
  \renewcommand{\arraystretch}{1.15}
  \caption{Reproduction configuration for each baseline method. All methods use \texttt{claude-sonnet-4.6} for skill generation and share the same SPARK teacher trajectories as input.}
  \label{tab:baseline_config}
  \begin{tabular}{@{} l p{10.5cm} @{}}
    \toprule
    \textbf{Method} & \textbf{Configuration} \\
    \midrule
    Trace2Skill
      & \textbf{Stage 1 (Step-level Lesson Extraction):} per-attempt lesson extraction; input truncation: commands 6\,000 chars, stdout 4\,000 chars; temperature 0.3, max tokens 2\,048. \textbf{Stage 2 (Lesson Aggregation + Skill Synthesis):} aggregate lessons across attempts with \texttt{[SUCCESS]}/\texttt{[FAIL]} labels; aggregation truncated to 12\,000 chars; temperature 0.5, max tokens 4\,096. \\
    \midrule
    AutoRefine
      & \textbf{Step 1 (Initial Skill):} generated from the first attempt; input truncation: commands 6\,000 chars, stdout 4\,000 chars; temperature 0.5, max tokens 4\,096. \textbf{Step 2 (Iterative Refinement):} each subsequent attempt refines the current skill document (truncated to 8\,000 chars); temperature 0.4, max tokens 4\,096. \\
    \midrule
    EvoSkill
      & \textbf{Population:} 4 candidates with diversity hints (workflow ordering, failure modes, environment setup, implementation details). \textbf{Generations:} 2; top-$k$: 2 parents retained per generation. \textbf{Fitness:} LLM-based scoring (completeness, specificity, actionability, accuracy, organization; 0--100); temperature 0.2. \textbf{Mutation:} temperature 0.6; \textbf{Crossover:} temperature 0.5. \\
    \bottomrule
  \end{tabular}
\end{table}

\paragraph{Results.}
\autoref{tab:baseline_results} reports the comparison. SPARK achieves the highest $\Delta$\,Pass Rate ($+$28.8\%) and $\Delta$\,MR ($+$0.291), outperforming the strongest baseline AutoRefine by $+$10.2\% in pass rate gain and $+$0.119 in mean reward gain. SPARK also exhibits the most favorable improvement-to-degradation ratio (21 improved vs.\ 3 degraded tasks), whereas AutoRefine improves 19 tasks but degrades 9. These results confirm that PDI-guided posterior distillation produces substantially stronger skills than methods that rely on lesson extraction, iterative refinement, or evolutionary search without environment-grounded trajectory analysis.

\begin{table}[h]
  \centering
  \small
  \setlength{\tabcolsep}{4pt}
  \renewcommand{\arraystretch}{1.15}
  \caption{Comparison of skill generation methods on SkillsBench. BL: baseline (no skill); Gen: with generated skill; MR: mean reward; Impr./Degr.: number of tasks improved/degraded by the generated skill relative to baseline. $\Delta$ values measure the gain attributable to the skill.}
  \label{tab:baseline_results}
  \begin{tabular}{@{} l c c c c c c c c @{}}
    \toprule
    \textbf{Method} & \textbf{BL Pass} & \textbf{BL MR} & \textbf{Gen Pass} & \textbf{Gen MR} & \textbf{$\Delta$ Pass} & \textbf{$\Delta$ MR} & \textbf{Impr.} & \textbf{Degr.} \\
    \midrule
    Trace2Skill & 16.9\% & 0.218 & 27.1\% & 0.348 & +10.2\% & +0.130 & 12 & 4 \\
    AutoRefine  & 22.0\% & 0.261 & 40.7\% & 0.434 & +18.6\% & +0.172 & 19 & 9 \\
    EvoSkill    & 20.3\% & 0.246 & 30.5\% & 0.353 & +10.2\% & +0.107 & 11 & 2 \\
    \textbf{SPARK (Ours)} & 22.0\% & 0.240 & \textbf{50.8\%} & \textbf{0.531} & \textbf{+28.8\%} & \textbf{+0.291} & \textbf{21} & \textbf{3} \\
    \bottomrule
  \end{tabular}
\end{table}

\section{External Benchmark: ALFWorld}
\label{app:alfworld}

To evaluate whether SPARK generalizes beyond terminal-based programming tasks, we conduct an external evaluation on ALFWorld~\citep{shridhar2021alfworld}, a text-based interactive household environment where agents issue natural-language commands to manipulate objects. ALFWorld differs fundamentally from SkillsBench: actions are discrete text commands rather than shell operations, feedback is environment state descriptions rather than program output, and success requires multi-step physical reasoning. This evaluation simultaneously validates both the SPARK skill generation pipeline and the PDI-guided refinement mechanism in an out-of-distribution setting.

\paragraph{Setup.}
We evaluate on the ALFWorld \texttt{eval\_out\_of\_distribution} split (134 games, 6 task types). The agent framework is ReAct (1-shot, per task type) with a maximum of 30 steps per game. \autoref{tab:alfworld_config} summarizes the evaluation configuration.

\begin{table}[h]
  \centering
  \small
  \setlength{\tabcolsep}{5pt}
  \renewcommand{\arraystretch}{1.15}
  \caption{ALFWorld evaluation configuration.}
  \label{tab:alfworld_config}
  \begin{tabular}{@{} l l @{}}
    \toprule
    \textbf{Parameter} & \textbf{Value} \\
    \midrule
    Benchmark & ALFWorld (eval\_out\_of\_distribution, 134 games) \\
    Task types & 6 (Pick \& Place, Clean \& Place, Heat \& Place, Cool \& Place, Examine, Pick Two) \\
    Teacher models & Claude Sonnet 4.5, GPT-5.5 \\
    Student model & Claude Haiku 4.5 \\
    Teacher exploration & Per teacher \\
    Student evaluation & 3 conditions (No Skill, +Skill w/o PDI, +Skill w/ PDI) $\times$ 2 teachers \\
    Agent framework & ReAct (1-shot, per task type) \\
    Max steps & 30 steps/game \\
    \bottomrule
  \end{tabular}
\end{table}

\paragraph{Skill generation and PDI refinement.}
For each teacher, we run 30 exploration games and collect trajectories. Skills are distilled per task type under two conditions: without PDI filtering and with PDI-guided trajectory selection. PDI scores are computed per trajectory; the PDI condition prioritizes high-PDI trajectories during distillation.

\paragraph{Results.}
\autoref{tab:alfworld_sonnet} and \autoref{tab:alfworld_gpt} report per-task-type success rates for the two teacher models. Under both teachers, SPARK-generated skills improve overall success rate over the no-skill baseline, and PDI-guided skills consistently outperform their non-PDI counterparts. With GPT-5.5 as teacher, PDI-refined skills achieve 40.0\% overall success (vs.\ 16.7\% baseline), with gains appearing even on task types where the teacher itself never succeeded during exploration (e.g., Pick \& Place, Cool \& Place), indicating that PDI-guided distillation can extract transferable procedural patterns from partial successes in related task types. For brevity, we omit \emph{Heat \& Place} and \emph{Pick Two} from the tables, since the baseline is already 0\% and all three conditions remain at 0\% (no change, neither gain nor drop).

\begin{table}[h]
  \centering
  \small
  \setlength{\tabcolsep}{5pt}
  \renewcommand{\arraystretch}{1.15}
  \caption{ALFWorld results with Claude Sonnet 4.5 as teacher (exploration success: 25\%). Student: Claude Haiku 4.5.}
  \label{tab:alfworld_sonnet}
  \begin{tabular}{@{} l c c c @{}}
    \toprule
    \textbf{Task Type} & \textbf{No Skill} & \textbf{+Skill (no PDI)} & \textbf{+Skill (PDI)} \\
    \midrule
    Pick \& Place   & 0.0\%  & 0.0\%  & 0.0\%  \\
    Clean \& Place  & 0.0\%  & 20.0\% & \textbf{40.0\%} \\
    Cool \& Place   & 0.0\%  & 0.0\%  & 0.0\%  \\
    Examine         & 71.4\% & 85.7\% & \textbf{100.0\%} \\
    \midrule
    \textbf{Overall} & 16.7\% & 23.3\% & \textbf{30.0\%} \\
    \bottomrule
  \end{tabular}
\end{table}

\begin{table}[h]
  \centering
  \small
  \setlength{\tabcolsep}{5pt}
  \renewcommand{\arraystretch}{1.15}
  \caption{ALFWorld results with GPT-5.5 as teacher (exploration success: 23\%). Student: Claude Haiku 4.5.}
  \label{tab:alfworld_gpt}
  \begin{tabular}{@{} l c c c @{}}
    \toprule
    \textbf{Task Type} & \textbf{No Skill} & \textbf{+Skill (no PDI)} & \textbf{+Skill (PDI)} \\
    \midrule
    Pick \& Place   & 0.0\%  & 0.0\%  & \textbf{25.0\%} \\
    Clean \& Place  & 0.0\%  & 20.0\% & \textbf{40.0\%} \\
    Cool \& Place   & 0.0\%  & 14.3\% & \textbf{28.5\%} \\
    Examine         & 71.4\% & \textbf{100.0\%} & \textbf{100.0\%} \\
    \midrule
    \textbf{Overall} & 16.7\% & 30.0\% & \textbf{40.0\%} \\
    \bottomrule
  \end{tabular}
\end{table}

\paragraph{Reproduction.}
All experiments use the ALFWorld Python package (\texttt{pip install alfworld}) with the default \texttt{eval\_out\_of\_distribution} split. Teacher exploration runs with 10 parallel workers. Skill distillation uses \texttt{claude-sonnet-4.5} with and without the PDI flag. Student evaluation runs across 3 conditions with 10 parallel workers per condition. Scripts are located in \texttt{extra/benchmark/scripts/}: \texttt{launch\_parallel.py} (teacher exploration), \texttt{skill\_gen.py} (distillation), \texttt{launch\_parallel\_eval.py} (student evaluation), and \texttt{generate\_tables.py} (result aggregation).

\section{Limitations and Scope}
\label{app:limitations}

Our study focuses on text- and terminal-based agent environments and validates SPARK on SkillsBench, 300 SPARK-generated task variants, and the out-of-distribution ALFWorld benchmark. Extending the same trajectory-level analysis to other agent modalities (e.g., GUI-based or multi-modal settings) is a natural direction that we leave for future work. PDI is intentionally instantiated as a simple, equal-weight linear composite of three interpretable signals, which we find to be a robust default across held-out folds (\appref{app:weight_sensitivity}); a fully learned or domain-adapted variant may be beneficial when a large, homogeneous task distribution is available.


\clearpage

\end{document}